%% file: main.tex
\pdfoutput=1 

\documentclass[11pt,a4paper]{article}
\usepackage{acl}
\usepackage{nert}
\usepackage{times}
\usepackage{float}
\usepackage{latexsym}
\usepackage{tabularx}
\usepackage{tabulary}
\usepackage{colortbl}
\usepackage{xcolor}
\definecolor{Gray}{gray}{0.92}
\newcolumntype{Y}{>{\centering\arraybackslash}X}
\usepackage[noabbrev,capitalise]{cleveref}

\input{macros}

\usepackage{color,soul}
\definecolor{lightred}{HTML}{FF6D6A}
\definecolor{lightgreen}{HTML}{77DD77}

\usepackage{todonotes}
\makeatletter
\newcommand*\iftodonotes{\if@todonotes@disabled\expandafter\@secondoftwo\else\expandafter\@firstoftwo\fi}
\makeatother

\newcommand{\note}[4][]{\todo[author=#2,color=#3,size=\scriptsize,fancyline,caption={},#1]{#4}}

\newcommand{\srcomment}[2][]{\note[#1]{SR}{yellow!70}{\small #2}}

\title{Image Retrieval from Contextual Descriptions}

\author{
    Benno Krojer\textsuperscript{1}~\;~
    Vaibhav Adlakha\textsuperscript{1}~\;~
    Vibhav Vineet\textsuperscript{3} \\
    \bf Yash Goyal\textsuperscript{4}~\;~
    Edoardo Ponti\textsuperscript{1}~\;~
    Siva Reddy\textsuperscript{1,2} \\
    \textsuperscript{1}Mila/McGill University~\;~
    \textsuperscript{2}Facebook CIFAR AI Chair \\
    \textsuperscript{3}Microsoft Research~\;~
    \textsuperscript{4}Samsung - SAIT AI Lab, Montreal\\
    \href{mailto:benno.krojer@mila.quebec}{\texttt{benno.krojer@mila.quebec}}~\:~
    \href{mailto:siva.reddy@mila.quebec}{\texttt{siva.reddy@mila.quebec}}
}

\begin{document}

\maketitle

\begin{abstract}
The ability to integrate context, including perceptual and temporal cues, plays a pivotal role in grounding the meaning of a linguistic utterance. In order to measure to what extent current vision-and-language models master this ability, we propose a new multimodal challenge, \datasetname{} (\datasetacronym{}). In particular, models are tasked with retrieving the correct image from a set of 10 minimally contrastive candidates based on a contextual description.
As such, each description contains only the details that help distinguish between images.
Because of this, descriptions tend to be complex in terms of syntax and discourse and require drawing pragmatic inferences. 
Images are sourced from both static pictures and video frames.
We benchmark several state-of-the-art models, including both cross-encoders such as ViLBERT and bi-encoders such as CLIP, on \datasetacronym{}.
Our results reveal that these models dramatically lag behind human performance: the best variant achieves an accuracy of 20.9 on video frames and 59.4 on static pictures, compared with 90.8 in humans.
Furthermore, we experiment with new model variants that are better equipped to incorporate visual and temporal context into their representations, which achieve modest gains. 
Our hope is that \datasetacronym{} will foster progress in grounded language understanding by encouraging models to focus on fine-grained visual differences.
We make code and dataset publicly available.\footnote{\href{https://github.com/McGill-NLP/imagecode}{https://github.com/McGill-NLP/imagecode}}
\end{abstract}

\section{Introduction}

\begin{figure}[h!]
\centering
      \begin{subfigure}[t]{0.49\columnwidth}
          \centering
 \includegraphics[width=\textwidth]{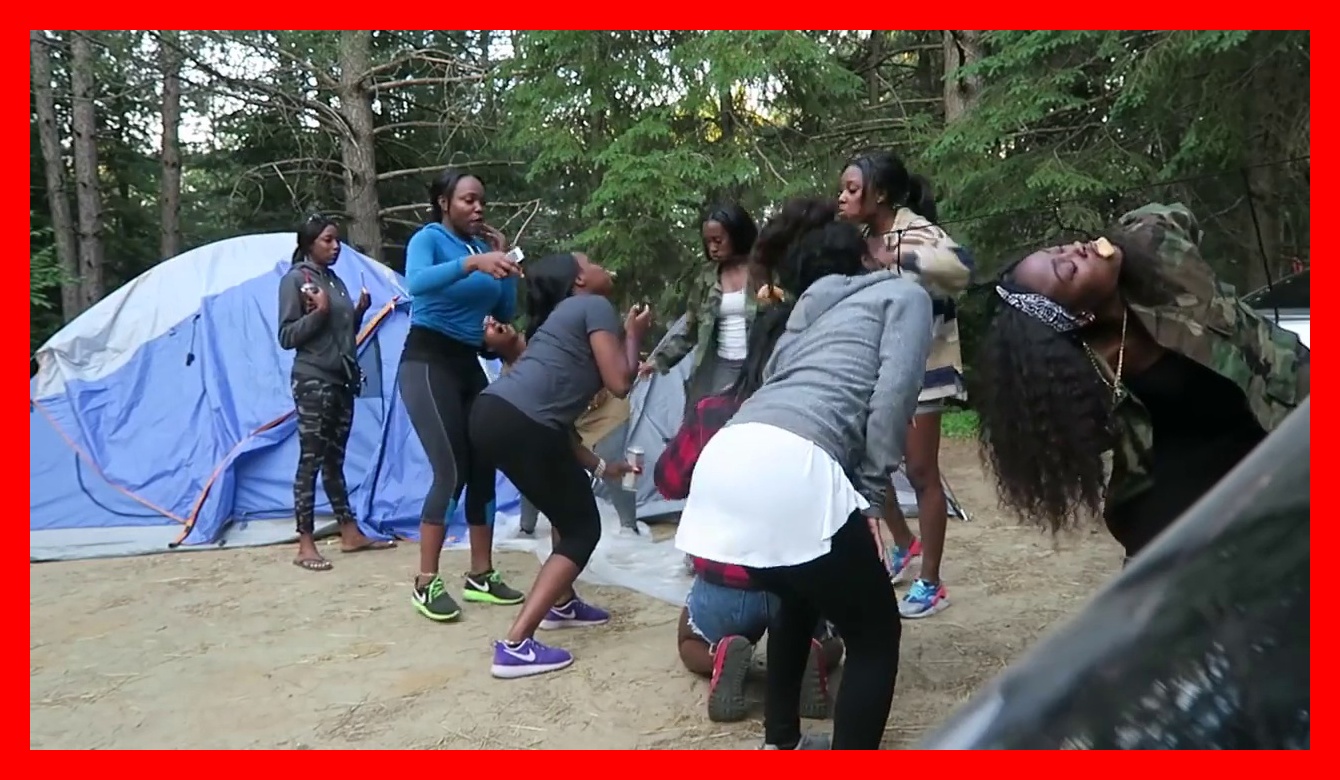}
          \caption{{Frame 1}}
          \label{subfig:frame3}
      \end{subfigure}
      \hfill
     \begin{subfigure}[t]{0.49\columnwidth}
         \centering
         \includegraphics[width=\textwidth]{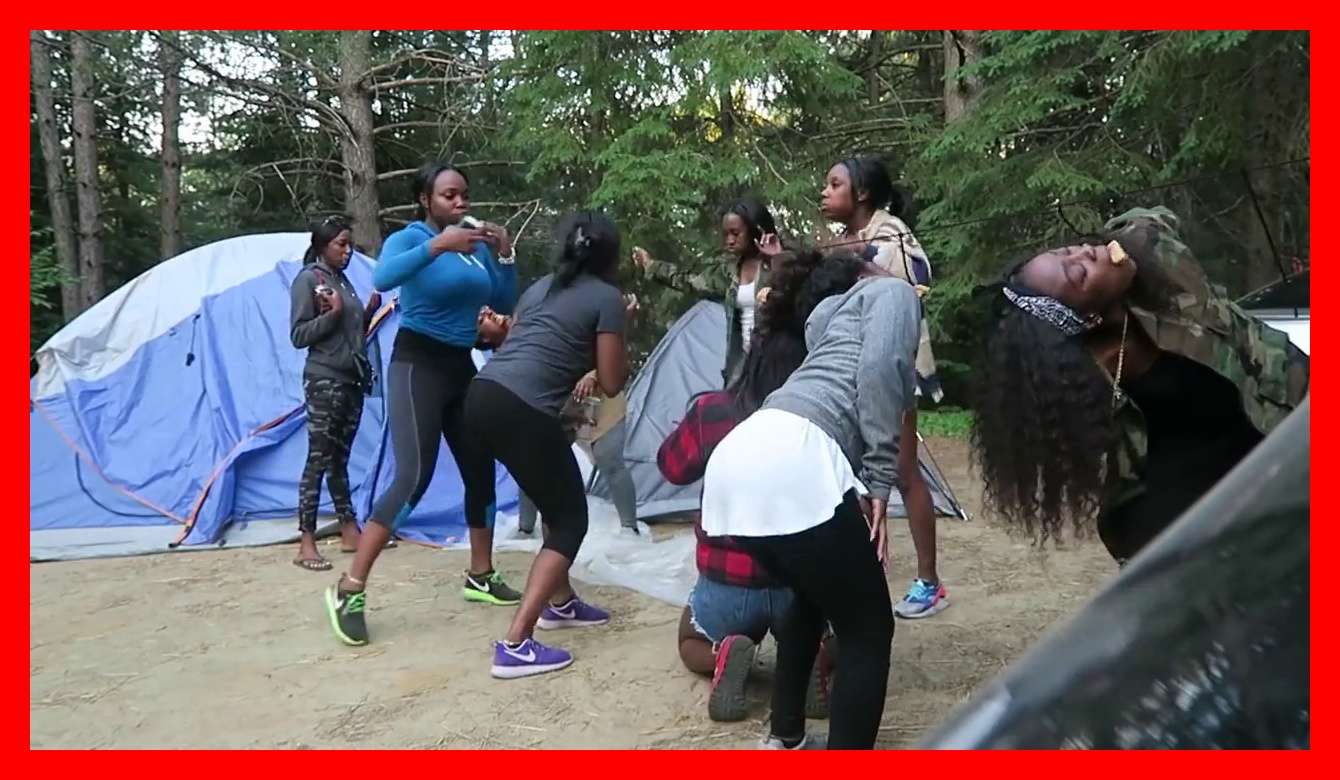}
         \caption{{Frame 2}}
         \label{subfig:frame4}
     \end{subfigure}

     \begin{subfigure}[t]{0.49\columnwidth}
         \centering
\includegraphics[width=\textwidth]{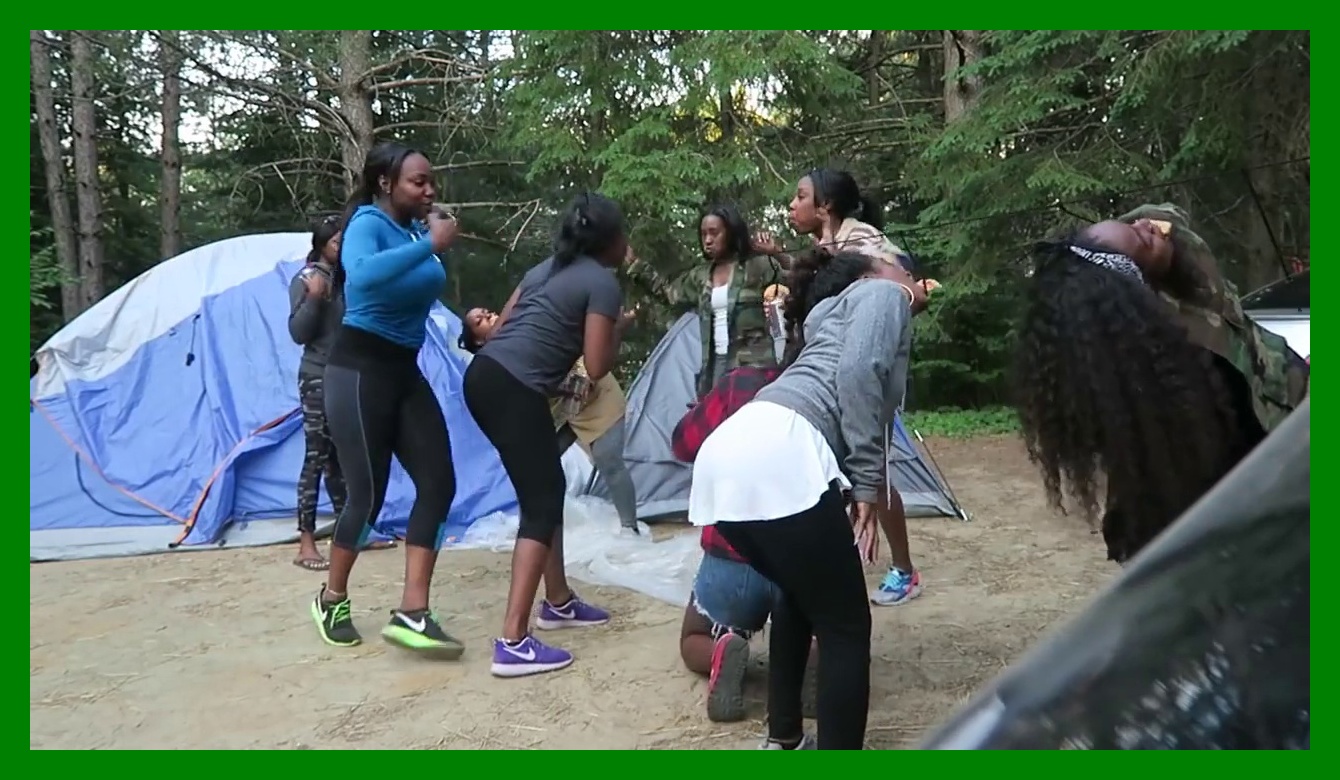}
         \caption{{Frame 3}}
         \label{subfig:frame5}
     \end{subfigure}
     \hfill
     \begin{subfigure}[t]{0.49\columnwidth}
         \centering
         \includegraphics[width=\textwidth]{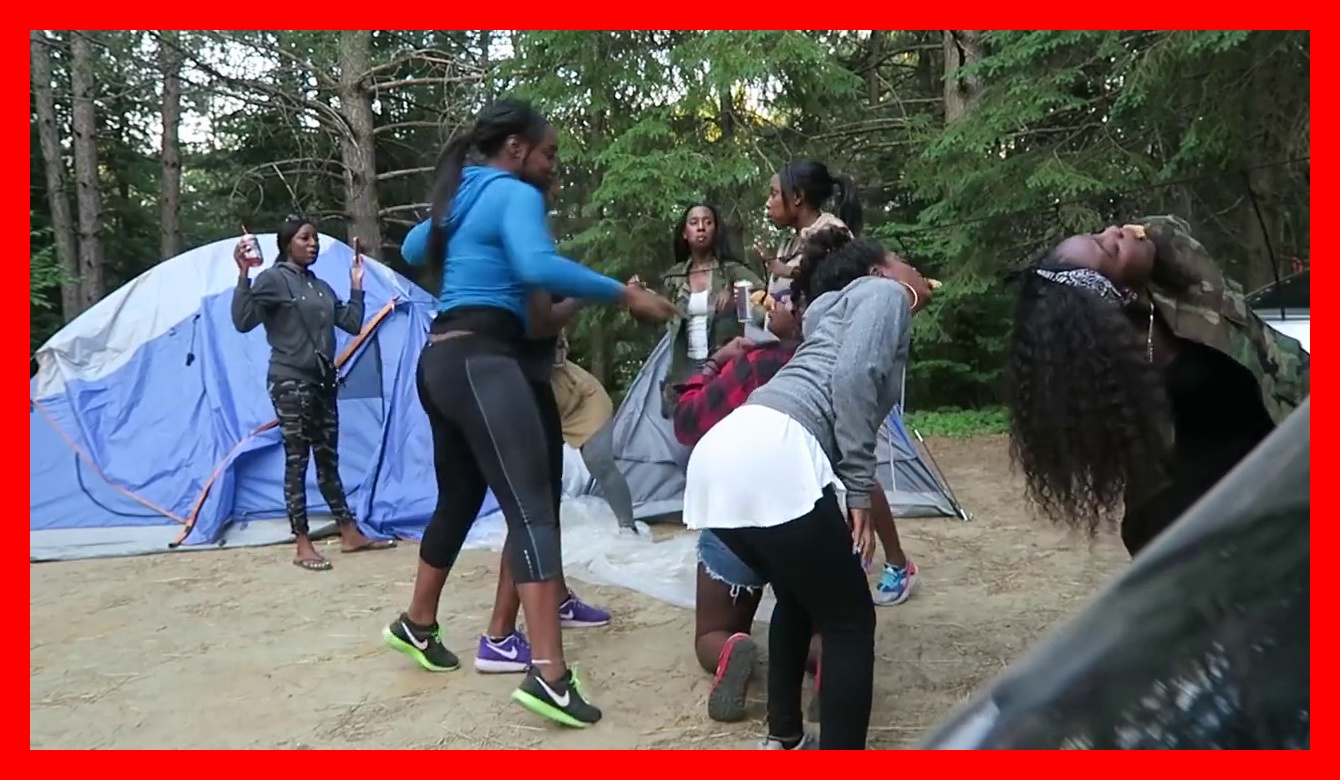}
         \caption{{Frame 4}}
         \label{subfig:frame6}
     \end{subfigure}
\caption{
An example of the new challenge, \datasetname{} (\datasetacronym{}):
\textit{``The girl in blue is to the left of the girl in the middle with the purple shoes. The girl in blue is not obscured in any way.''}
Frames 5--10 are left out for simplicity's sake. The target image, frame 3, is in green, whereas the incorrect frames are in red.
}
\label{fig:cidexample}
\end{figure}

Natural languages are highly contextual \citep{fodor2001language}: for a listener, recovering the speaker's \textit{intended} meaning requires integrating information from different streams, such as grounding in perception \citep{pecher2005grounding}, shared world knowledge, and temporal reasoning \citep{wilson1998pragmatics}. These processes, more generally, fall under the umbrella term of \textit{pragmatics} \citep{grice1957meaning}. Despite recent progress in multimodal systems, it remains unclear to which extent they can handle settings where context plays a major role, such as in real-world communication.

To this end, we present a new challenge that requires multimodal models to leverage context to retrieve images from text. In particular, given a contextual description and a set of minimally contrastive candidate images, i.e.\ differing only in some details,
the model has to retrieve the target image. In order to discriminate between similar images, human annotators naturally produce highly nuanced and grammatically complex descriptions. An example of our new challenging dataset, \datasetname{} (\datasetacronym{}), is shown in \cref{fig:cidexample}.

During the data collection process, sets of similar images are selected among static pictures from Open Images \cite{OpenImages} and (a larger portion) among video frames from diverse domains. Including both types of images allows for diversifying the dataset while representing different degrees of visual similarity within each set.
Next, we crowdsource a \textit{contextual} description of a target image (presented together with the rest of the set) that contains only differences relevant for retrieval. 
After a filtering phase involving human retrievers, we obtain a large-scale dataset with 94,020 images and 21,202 descriptions associated with image sets of size 10.


As a result of this annotation protocol, successfully completing the task requires models to integrate several kinds of context: \textbf{i}) the image set, as the descriptions often only make sense in the context of several other images and are not suitable as stand-alone captions. In fact, aspects of the image that are very salient and that therefore would normally be emphasized are not useful in our proposed task. Instead, the focus of our descriptions are fine-grained details that help discriminate between images (see \cref{fig:cidexample}); \textbf{ii}) the speaker's intention. Due to their high degree of image similarity, contextual descriptions may be literally true for multiple images; however, once the speaker's intention is taken into account, the correct image can be determined by virtue of pragmatics, i.e. Grice's maxim of quality \footnote{Note: While we do not model pragmatics explicitly in our baselines, we find that the \datasetacronym{} contains many examples suitable for pragmatic modeling} (see \cref{fig:contextuality}, \cref{bread}); \textbf{iii}) temporal sequences: for video frames temporal reasoning is also required to compare different moments of an unfolding event.

On our new dataset \datasetacronym{}, we benchmark a series of vision-and-language models
that achieve state-of-the-art performance on other multimodal tasks, specifically ViLBERT \citep{lu_vilbert_2019} and UNITER \cite{vedaldi_uniter_2020} as two cross-encoder variants and CLIP as a strong bi-encoder \citep{radford2021learning}.
We report several findings. First, accuracy on static images is vastly superior than on video frames. Therefore, the degree of similarity among the candidate images has an overwhelming impact on retrieval performance.  Second, all state-of-the-art models generally struggle with image retrieval from contextual descriptions, whereas humans consistently achieve high accuracy.

Hence, we propose model variants capable of better taking context into account: \textbf{i}) once an image-description pair is encoded, we refine this representation by attending to the other images in the set; \textbf{ii}) we augment image encodings with temporal embeddings. Based on our results, models take advantage of this additional information fruitfully but only to a limited degree.

Because of its challenging nature, due to the minimally contrastive images and complex descriptions, we believe that \datasetacronym{} will help make visio-linguistic models more context-aware and sensitive to fine-grained details.
\section{Related Work}
\label{sec:RW}

%
There is a long tradition of grounding language understanding on single images, in the form of visual question answering \cite{goyal2017making, hudson2019gqa}, visual dialogue \cite{de2017guesswhat, das2017visual}, or visual entailment \cite{xie2019visual}.
Recently, more and more focus has been directed to settings where the visual context consists of multiple images, either conventional static pictures \cite{vedantam_context-aware_2017, hexiang2018bison,suhr_corpus_2019,forbes_neural_2019,hendricks_probing_2021,yan_l2c_2021,hosseinzadeh2021image, bogin2021covr,liu-etal-2021-visually}, or video frames \cite{jhamtani_learning_2018, Bansal2020VisualQA}.
While many of these benchmarks involve just two images, COVR \cite{bogin2021covr} and ISVQA \cite{Bansal2020VisualQA} provide more images, similar to our sets of 10 images.

ISVQA and Spot-the-diff \cite{jhamtani_learning_2018} are most similar to our dataset, \datasetacronym{}.
ISVQA is based on several video frames that are synthetic and cover a restricted domain, with short questions for Visual Question Answering.
Spot-the-diff provides two frames from surveillance video cameras and descriptions of all their differences.
\datasetacronym{} is unique as a) we cover a wider range of domains; b) we construct image sets that are maximally similar while being distinguishable through natural language (\cref{sec:datacollection}) and c) we limit descriptions to \textit{relevant} differences.
This results in (a) diverse, (b) complex and (c) pragmatically informative descriptions.

We do not claim to explicitly model pragmatics in this paper, i.e. with \textit{Rational Speech Acts} \cite{goodman2016pragmatic}. Instead we present a dataset that is naturally suitable for pragmatic reasoning \cite{andreas_reasoning_2016, cohn2018pragmatically} as a listener has to consider the context, assume a Gricean speaker and resolve ambiguities resulting from nuanced differences.
The reasoning in our task and data collection is therefore also similar to ReferItGame and subsequent work \cite{kazemzadeh2014referitgame, reg} where one crowdworker generates a referring expressing for an object in a single image and another worker picks an object based on the expression.

\section{Data Collection}
\label{sec:datacollection}

Our data collection involves two steps with a human describer and retriever.
The describer is given a set of 10 highly similar images $S=[I_1,I_2,...,I_{10}]$, one of them marked as the target image $I_t$, and has to write a description $D$ that clearly distinguishes $I_t$ from the other distractor images.
In the second step, the retriever is given the same 10 images and the description from the first step and has to identify the target image based on the description.
$S$ and $D$ are only added to our dataset if the retrieval is successful. 

Below, we outline the main stages of data collection: first, the collection of similar, contrastive images in \cref{ssec:imagecollection}. Then, the crowdsourcing of contextual descriptions in \cref{ssec:crowddesc} and validation of the examples via image retrieval (\cref{ssec:crowdretr}). The final \datasetacronym{} dataset consists of 94,020 images (partitioned into 9,402 sets) and 21,202 contextual descriptions (16,594 in the train split, 2,302 and 2,306 in the validation and test split respectively).



\subsection{Collecting Similar Images}
\label{ssec:imagecollection}
In the first stage, we collect sets of images that are highly similar but still distinguishable from each other by a human. 
To quantitatively measure the pairwise similarity of two images, we compute the Euclidean distance between their encodings extracted from a pre-trained CLIP model \citep{radford2021learning}.\footnote{We also experimented with ResNet-50 features, but we found CLIP results to be more similar to that of humans in preliminary experiments. 
} To study the effect of different degrees of similarity, further variegate our dataset, and enable temporal reasoning, we source our candidate images from collections of static pictures as well as videos, as detailed below.

\paragraph{Static Pictures.}
We obtain image sets from one of the largest repositories of static pictures, the Open Images Dataset V6 \citep{OpenImages}, containing 1.74M images. For each image, we retrieve the 9 closest images from the training set based on their CLIP encodings. We then randomly sample 4,845 of these image sets. 

\paragraph{Video Frames.}
As sources for our video frames, we use \textbf{i}) Video-Storytelling \citep{li2019video}, covering social events (wedding, birthday, Christmas, camping); \textbf{ii}) general-domain MSR-VTT \cite{xu2016msr}; and \textbf{iii}) YouCook \citep{das2013thousand}, covering cooking events.
We choose these datasets as they contain publicly available and general-purpose videos (not specific to downstream tasks). We retain the original splits for train, validation, and test.

To obtain disjoint sets of 10 similar frames, we first segment the videos into smaller scenes (also known as shots) via the scene detection functionality of \texttt{ffmpeg} \cite{tomar2006converting}. Then, for each scene, we add its first frame to the set of selected images. We then iterate over every following frame and add it to the set if its pairwise Euclidean distance with each of the previously selected frames is larger than a threshold.\footnote{The distance threshold was manually chosen as 0.35 based on qualitative results.} Once the set contains 10 images, we reiterate the procedure for a new set. If the scene ends and the current set contains less than 10 images, the set is discarded.

During this process, we additionally remove frames that \textbf{i}) are too blurry, i.e.\ their BRISQUE score \cite{mittal2012no} is larger than 0.65; or \textbf{ii}) contain too much text, which is detected with the OCR tool Tesseract \citep{smith2007overview}.\footnote{The rationale of the second criterion is to prevent workers from focusing on the overlaid text rather than image content.}
We use all of YouCook's image sets and (due to cost constraints) randomly sample image sets from Video-Storytelling and MSR-VTT for crowdsourcing (cf.\ \cref{tab:dataset_size}). 
We remark that image sets are further filtered at the final stage of annotation (\cref{ssec:crowdretr}).


\begin{table}[t]
    \centering
    \begin{tabular}{c r r}
    \toprule
    Dataset & After \S\ref{ssec:imagecollection} & After \S\ref{ssec:crowdretr} \\
    \midrule
    MSR-VTT & 11,643 & 8,045 \\
    Video-Storytelling & 11,459 & 8,153 \\
    YouCook & 894 & 588 \\
    Open Images & 4,845 & 4,416 \\
    \bottomrule
    \end{tabular}
    \caption{
    Number of descriptions from each source of images at different stages of the annotation process.}
    \label{tab:dataset_size}
    \vspace{-2mm}
\end{table}

\begin{table*}[t]
   \footnotesize
    \centering
    \begin{tabularx}{\textwidth}{Y | rrr | p{6cm} | p{3.2cm}}
    \toprule
    \textbf{Phenomenon} & \textbf{all} & \textbf{videos} & \textbf{static} & \textbf{Example from \datasetacronym{}} & \textbf{Definition} \\
    & \% & \% & \% & & \\
    \midrule
    \multirow{2}{*}{\underline{Context}} & \multirow{2}{*}{47.3} & \multirow{2}{*}{\textbf{57.3}} & \multirow{2}{*}{6.6} & \multirow{2}{*}{\cref{fig:contextuality}} & \scriptsize{Visual context or pragmatic inference required.}
    \\ \hline
    
    \multirow{2}{*}{Temporal} & \multirow{2}{*}{15.0} & \multirow{2}{*}{\textbf{18.5}} & \multirow{2}{*}{4.1} & \textit{A smiling boy just \textbf{begins to} look towards the dog.} & \scriptsize{Temporal markers (e.g., \textit{after}) and verbs (e.g., \textit{starts})} \\ \hline
    
    \multirow{2}{*}{Quantities} & \multirow{2}{*}{48.5} & \multirow{2}{*}{47.7} & \multirow{2}{*}{\textbf{51.0}} & \textit{There is an \textbf{equal amount} of yellow and white between \textbf{both} hands.} & \multirow{2}{*}{---} \\ \hline
    
    \multirow{2}{*}{Spatial Relations} & \multirow{2}{*}{70.5} & \multirow{2}{*}{\textbf{72.2}} & \multirow{2}{*}{65.3} & \textit{The cloud on \textbf{top left side} of box only has \textbf{half of} it showing.} & \multirow{2}{*}{---} \\ \hline
    
    \multirow{2}{*}{\underline{Negation}} & \multirow{2}{*}{17.9} & \multirow{2}{*}{\textbf{20.7}} & \multirow{2}{*}{6.1} & \textit{The spoon is at the top right corner, it is \textbf{not} moving any of the food.} & \multirow{2}{*}{---} \\ \hline
    
    \underline{Visibility} / \underline{Occlusion} & \multirow{2}{*}{45.5} & \multirow{2}{*}{\textbf{54.5}} & \multirow{2}{*}{8.6} & \textit{The flowers the woman in the teal strapless dress is carrying are \textbf{completely obscured} by the man in the black shirt's head.} & \scriptsize{An entity is covered or partially outside of the image.} \\ \hline
    
    \multirow{2}{*}{\underline{Nuances}} & \multirow{2}{*}{26.3} & \multirow{2}{*}{\textbf{31.6}} & \multirow{2}{*}{5.1} & \textit{There is the \textbf{slightest of openings} to see the end of the bridge through the obstruction.} & \scriptsize{Description grounded on small patch of pixels or very non-salient aspects.} \\ \hline
    
    \multirow{2}{*}{Co-reference} & \multirow{2}{*}{41.5} & \multirow{2}{*}{\textbf{42.4}} & \multirow{2}{*}{38.8} & \textit{The cloud on top left side of box only has half of \textbf{it} showing.} & \multirow{2}{*}{---} \\ \hline
    
    \multirow{2}{*}{Meta Properties} & \multirow{2}{*}{12.0} & \multirow{2}{*}{\textbf{13.9}} & \multirow{2}{*}{6.1} & \textit{\textbf{Bright shot} of a girl and boy standing up straight. Her eyes are closed.} & \scriptsize{Blurriness, brightness, overlays, and transitions of frames.} \\ 
    \bottomrule
    
    
    \end{tabularx}
    \caption{Distribution of challenging phenomena in \datasetacronym{} based on 200 (or 1000 if underlined) manually annotated examples. 
    }
    \label{tab:phenomena}
\end{table*}

\subsection{Crowdsourcing Contextual Descriptions}
\label{ssec:crowddesc}

After creating sets of highly-similar images in \cref{ssec:imagecollection}, we request annotators from 
Amazon Mechanical Turk (AMT) to write contextual descriptions for each target image in a set. 
Each round, a set of images is presented in random order for static pictures and respecting temporal order for video frames. This encourages annotators to take the dynamics of the event into account.
We then (randomly) select 3 target images per set, and ask annotators to produce a description that discriminates them from the other images in the set. To encourage pragmatic reasoning, we do not ask for \textit{all} the differences (just those sufficient for retrieval) and do not allow explicit mentions of other images (see \cref{fig:contextuality}). 
We select high-quality annotators according to criteria in \cref{app:anncrit} and assign partly disjoint sets of annotators to train and test in order to avoid annotator bias \citep{geva_are_2019}.\footnote{For further details on crowdsourcing instructions, analysis of annotator bias and the AMT interface, please refer to \cref{app:bias} and \cref{app:interface}.}



\subsection{Human Validation via Image Retrieval}
\label{ssec:crowdretr}

Finally, we validate the annotation crowdsourced in \cref{ssec:crowddesc} by asking AMT workers to retrieve the correct target image from a set given its contextual description. For the final dataset, we retained only the examples that i) were retrieved successfully in the training set by a single worker or ii) were retrieved successfully by \textit{at least} 2 out of 3 workers in the validation and test sets. As a consequence, we filtered out 26.5\% of the contextual descriptions generated in \cref{ssec:crowddesc}. \Cref{tab:dataset_size}
compares the number of examples retained at each stage throughout the dataset creation.\footnote{Again, the set of workers validating train and test sets were partly disjoint to avoid annotator bias.}



\section{Data Analysis}
\label{sec:data_analysis}

\begin{figure*}
\centering
     \begin{subfigure}[t]{0.6\columnwidth}
         \centering
         \includegraphics[width=\textwidth]{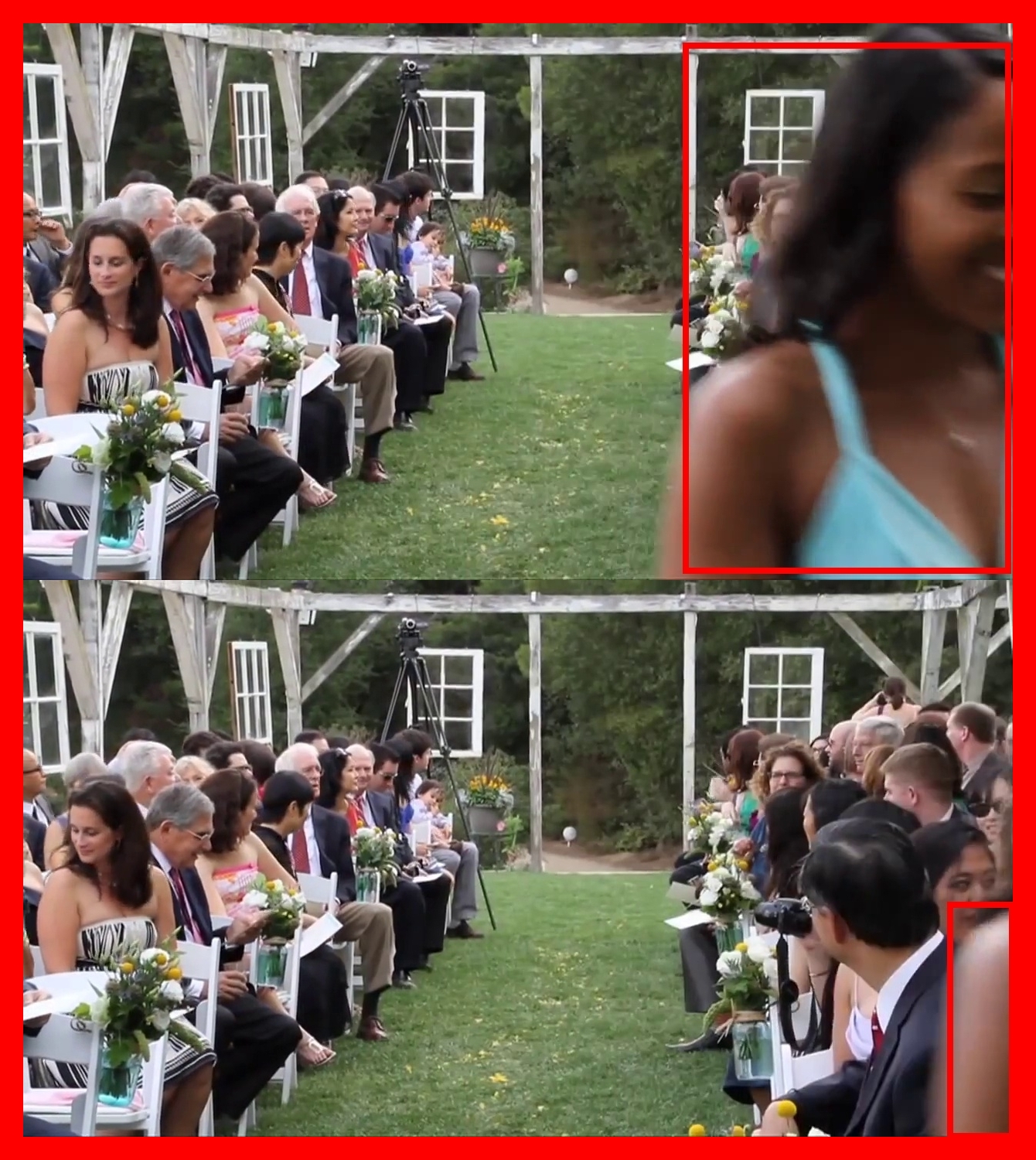}
         \caption{{Frame 1 \& Frame 2}}
         \label{subfig:framea}
     \end{subfigure}
     \hfill
     \hfill
     \begin{subfigure}[t]{0.8\columnwidth}
         \centering
         \includegraphics[width=\textwidth]{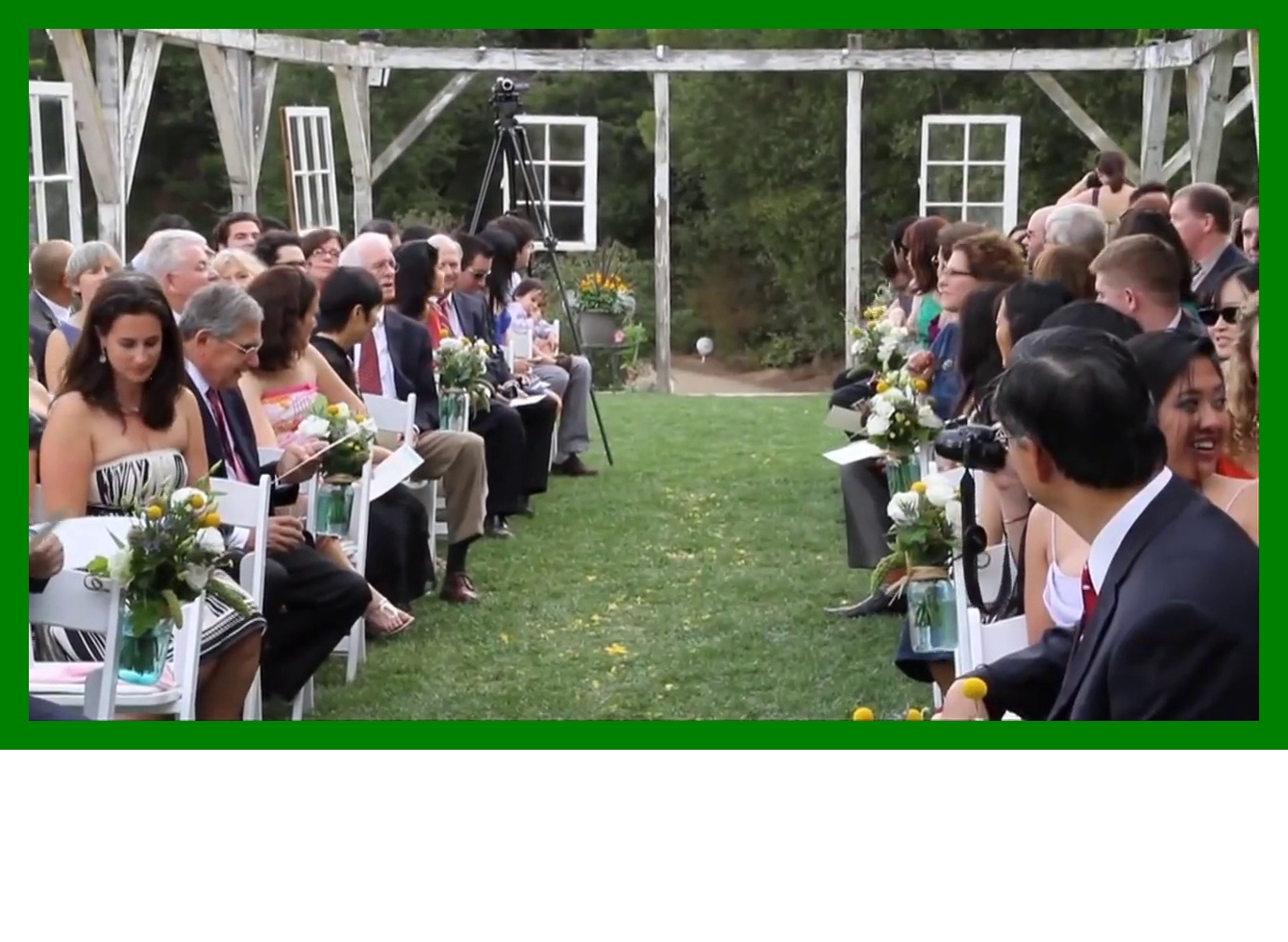}
         \caption{{Frame 3}}
         \label{subfig:frameb}
     \end{subfigure}
     \hfill
     \begin{subfigure}[t]{0.6\columnwidth}
         \centering
         \includegraphics[width=\textwidth]{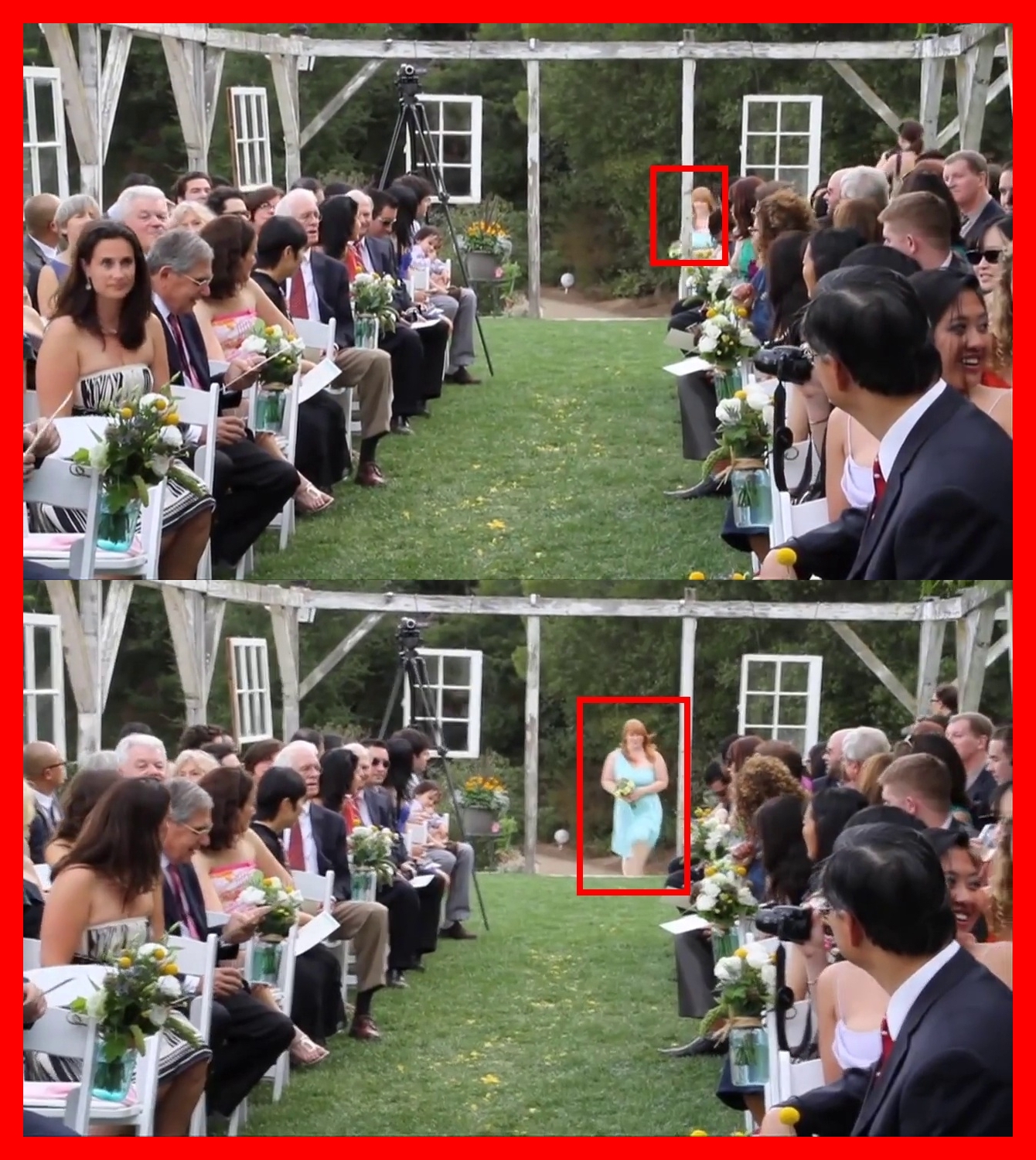}
         \caption{{Frame 4 \& Frame 5}}
         \label{subfig:framec}
     \end{subfigure}
     \hfill
\caption{
An example with description: 
``\textit{No bridesmaid visible at all.}''. Visual context is necessary to identify the correct target image, by cross-referencing the portions of images with bridesmaids (red boxes).
}
\label{fig:contextuality}
\end{figure*}

\subsection{Human Accuracy and Agreement}

To quantify the reliability of the process outlined in \cref{sec:datacollection}, we report the inter-annotator agreement on our final dataset in \cref{tab:humans}. We use Krippendorff's $\alpha$ as a metric (the higher the better), which accounts for incomplete data, since the number of annotators per example is not fixed. We treat the index of the target image either as a nominal variable for static images or as an ordinal variable for video frames. In both cases, we find a high degree of agreement. Moreover, in \cref{tab:humans}, we also report human accuracy-- the percentage of times an annotator retrieved the correct target image from a contextual description (as described in \cref{ssec:crowdretr}). This provides an upper ceiling for the model performances (see \cref{sec:results}).


\begin{table}[t]
    \centering\small
    \begin{tabular}{c r r}
    \toprule
    Metric & val & test \\
    \midrule
    Human Accuracy & 90.9 & 90.8 \\
    Krippendorff's $\alpha$ (nominal) & .797 & .795 \\
    Krippendorff's $\alpha$ (interval) & .872 & .869 \\
    \bottomrule
    \end{tabular}
    \caption{Human performance (accuracy) and inter-annotator agreement (Krippendorff's $\alpha$) on the validation and test splits of \datasetacronym{}.}
    \label{tab:humans}
\end{table}

\begin{table}[t]
    \centering\small
    \begin{tabularx}{\columnwidth}{lrrr}
    \toprule
    & ours & {NLVR2} & {Spot-the-diff} \\
    \midrule
    Average length & 23.3 & 15.3 & 10.6 \\
    Word types & 6,916 & 6,602 & 2,282 \\
    Average tree depth & 5.1 & 4.8 & 4.3 \\
    Average sentences & 1.6 & 1.0 & 1.0 \\
    \bottomrule
    \end{tabularx}
    \caption{Comparison of the text statistics of \datasetacronym{} with other vision-and-language datasets.}
    \label{tab:dataset_stats}
\end{table}

\subsection{Language Statistics}
In \cref{tab:dataset_stats}, we measure a series of statistics of the descriptions collected for \datasetacronym{} and compare them with other vision-and-language datasets with multiple naturalistic images (cf.\ \cref{sec:RW}), such as NLVR2 \citep{suhr_corpus_2019} and Spot-the-diff \citep{jhamtani-berg-kirkpatrick-2018-learning}.\footnote{For comparability, we measured the statistics for all the datasets with the same tools.} In particular, we count the average description length, the number of distinct word types, the average dependency tree depth of each sentence,\footnote{We use spaCy \cite{spacy2} as a parser.} and the average number of sentences per description. Based on these metrics, we find evidence that \datasetacronym{}'s descriptions are longer and more syntactically complex than in the other datasets. Moreover, they include multiple sentences (11.8\% of examples have 3 or more).



\subsection{Vision Statistics}
\label{ssec:visionstats}
By calculating the average pairwise Euclidean distance between CLIP-based encodings of images in the same set, we find that video frames are more similar than static pictures -- as expected -- by a factor of 1.13.
Moreover, we find that descriptions of video frames mention human body parts (72.1\%) more often than static pictures (30.2\%). On the other hand, names of colors appear in descriptions of static pictures (61.4\%) more frequently than video frames (33.6\%).\footnote{We calculated these percentages based on a list of 171 body parts in English collected by \citet{annika_tjuka_2021_5572303} and a list of colors in English from
\href{https://games4esl.com/list-of-colors-in-english}{games4esl.com}.}
Thus, annotators resort to different strategies to discriminate between different types of image sets, focusing on the aspects that vary the most.


\subsection{Challenging Phenomena}

Finally, we identify 9 interesting and challenging phenomena in \datasetacronym{} and annotate whether they are present in 200 examples from the validation set. We provide the definition of each phenomenon, its frequency, and an illustrative example in \cref{tab:phenomena}. An example for each phenomena is given in \cref{sec:app-phenomena}. For 4 of these phenomena unique to \datasetacronym{}, we further annotated 800 examples for the purpose of error analysis in \cref{sec:results}. 
Inspecting these examples, we find a high number of cases where the visual context (47.0\%) is required to complete the task. For instance, consider \cref{fig:contextuality}:
the description \textit{``\textit{No bridesmaid visible at all.}''} requires a retriever to resolve the co-references of the entities in 5 frames.
In particular, the body parts of the bridesmaids (red boxes) visible in frames 2 and 4 would not be identifiable as such without frame 1 and 5, respectively (where they appear with matching dresses and flowers in their hands).
A common example we find in the data are "gradable" scenarios, i.e. ``\textit{The person is looking down}'' might be semantically true for more than one image but it fits best to the image where the person is looking down the most.


Another group of phenomena characteristic for \datasetacronym{} originates from its minimally contrastive setup:
annotators might focus on how an event unfolds over time (\textit{temporal context}), on what is missing in a specific frame but visible in the others (\textit{negation}), on what moved out of frame (\textit{visibility / occlusion}), or on small regions and patches of pixels (\textit{nuances}).
Importantly, these phenomena are less prominent in static pictures than in video frames (cf.\ \cref{tab:phenomena}). 

\section{Methods}
\label{sec:methods}

\begin{figure*}[t]
    \centering
    \includegraphics[width=\textwidth]{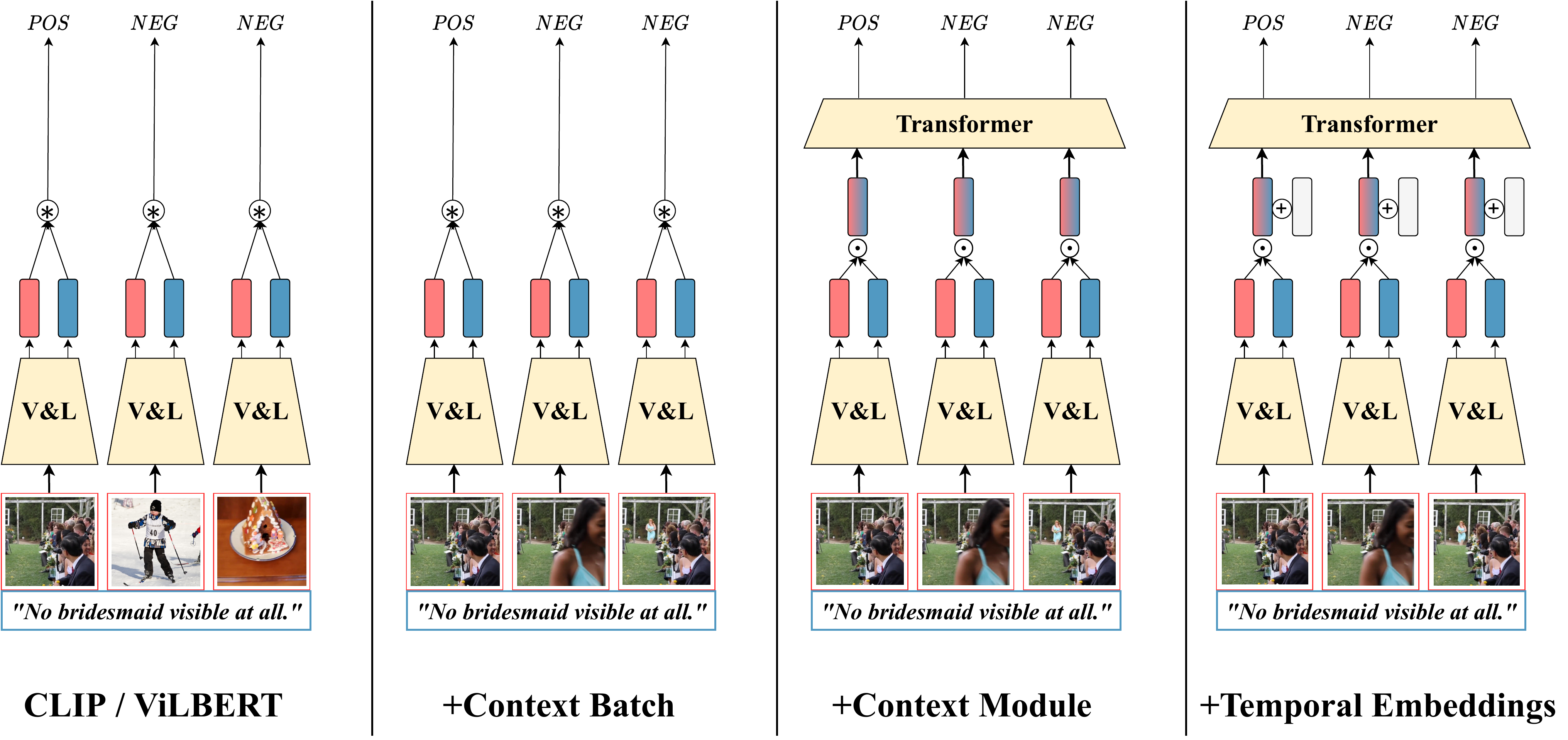}
    \caption{Models with increasing levels of context integration: see \cref{sec:methods} for more details. In the figure, we colour visual embeddings in red, text embeddings in blue, and positional embeddings in grey. \textit{POS} is the score for the target image and \textit{NEG} for the other candidates. $\oast$ represents dot product for CLIP and element-wise multiplication followed by a linear layer for ViLBERT/UNITER. $\odot$ represents element-wise multiplication. For ease of exposition, we show 3 images instead of 10.}
    \label{fig:models}
\end{figure*}

\subsection{Baselines}
In order to assess whether vision-and-language models can retrieve the correct image from a contextual description on a par with humans, we benchmark three state-of-the-art models that represent three main families of multimodal architectures \citep{bugliarello_multimodal_2021, miech_thinking_2021}: \textbf{i}) ViLBERT, a cross-encoder where language and vision streams can interact via cross-attention at intermediate layers \citep{lu_vilbert_2019}; \textbf{ii} UNITER, a single-stream encoder where language and vision tokens are concatenated as inputs and processed with a single Transformer \cite{vedaldi_uniter_2020}; \textbf{iii}) CLIP, a bi-encoder where language and vision streams are independent \citep{radford2021learning}. It is worth noting that ViLBERT and UNITER are more expressive due to their architecture, whereas CLIP boasts a higher parameter count, is pre-trained on a larger dataset and uses a contrastive objective.


We evaluate these models under two different regimes: i) \textit{zero-shot} inference, where pre-trained models are deployed on the \datasetacronym{} test set directly; and ii) \textit{fine-tuning}, where the models are refined on the full training set before evaluation. We cast the training objective as binary classification for ViLBERT and as 10-class classification for CLIP.\footnote{We found this solution to work better for each model in practice, which is justified by their different pre-training objectives.} Crucially, in both cases, positive and negative examples during training are sampled at random independently from the image set they belong to (see the first column of \Cref{fig:models}). 
Thus, the visual context of the other images in a set is only indirectly accessible at inference time, where the image with the highest probability is predicted.

\subsection{Integrating Context into Vision-and-Language Models}

For the fine-tuning regime, we further investigate some modifications in the training setup and model  architecture that facilitate the integration of visual and temporal context into the model. First, we use an alternative objective where all three models are trained on 10-class classification, but the 1 positive and 9 negatives are sourced from the same image set. The consequence of including positive and negative examples from the same image set in the same mini-batch is providing a wider visual context. We refer to this variant as \contextbatch{} (second column of \Cref{fig:models}).

This setup only conveys the visual context as a weak signal, since the model has no chance to directly compare the images in the same set. Hence, we experiment with enhancing the architecture of vision-and-language models with a mechanism inspired by \citet{bogin2021covr}. In particular, given an encoder (CLIP, ViLBERT or UNITER), we obtain the representations of a contextual description $\xx_L \in \mathbb{R}^e$ (where $e$ is the model hidden size) and of the images in a set $(\xx_V^{(1)}, \dots, \xx_V^{(10)}), \xx_V^{(i)} \in \mathbb{R}^e$ from their final layer.\footnote{We use the \textsc{CLS} tokens for UNITER/ViLBERT.}
Then, we create a series of multimodal embeddings via element-wise multiplication: $\mm = (\xx_L \odot \xx_V^{(1)}, \dots, \xx_L \odot \xx_V^{(10)})$. Finally, we feed these to a $l$-layer Transformer $\mathrm{Tf}: \mathbb{R}^{10\times e} \rightarrow \mathbb{R}^{10\times e}$ to obtain context-aware multimodal embeddings $(\mathrm{Tf}(\mm)_1, \dots, \mathrm{Tf}(\mm)_{10})$. Since each description--image pair can now attend on the others in a set, the model can fully exploit the visual context. We obtain the score for the $i$-th pair through a linear classifier head $W \in \mathbb{R}^{1 \times e}$. The target image is predicted as
\begin{equation} \label{eq:1}
    \argmax_i \; \mathrm{softmax}\left[W \left(\mathrm{Tf}(\mm)_i + \mm^{(i)}\right) \right]
\end{equation}
Note that we add a highway layer from the input to the output of the Transformer. We label this model variant \contextmodule{}.


Finally, in addition to visual context, we make models aware of the temporal context too, as shown in the fourth column of \Cref{fig:models}. 
For video-based examples only, the multimodal embeddings of each description-image pair are summed with a learnable positional embedding $\vt \in \mathbb{R}^e$ that reflects the temporal order of the frames.\footnote{In the examples with static pictures, no temporal embedding is added.} Thus, $\mm = (\xx_L \odot \xx_V^{(1)} \oplus \vt^{(1)}, \dots, \xx_L \odot \xx_V^{(10)} \oplus \vt^{(10)})$. 
Multimodal embeddings are then fed to a Transformer as above. We label this variant encapsulating both visual and temporal context \temporalemb{}.



\subsection{Experimental Setup}
\label{ssec:expsetup} 
For all CLIP experiments, we use a pre-trained model with the vision backbone \textsc{ViT-B/16}\footnote{\url{https://github.com/openai/CLIP}}. We train the full models with a batch size of 360 examples (i.e., 36 image sets) for CLIP and 150 examples for ViLBERT/UNITER. We perform early stopping based on the validation accuracy with a maximum of 30 epochs. In the variants that adopt the base version of a model, we select a learning rate of $4 \cdot 10^{-6}$ for CLIP, $5 \cdot 10^{-6}$ for ViLBERT, $4 \cdot 10^{-5}$ for ViLBERT\contextbatch{}, $8 \cdot 10^{-6}$ for UNITER, and $7 \cdot 10^{-6}$ for UNITER+\contextbatch{}. We find these values via hyper-parameter search on the range $[10^{-4}, 10^{-7}]$.

For CLIP variants that modify the model architecture, we adopt the following setup: first, we fine-tune the full model in the \contextbatch{} regime as detailed above. Afterwards, we freeze the encoder parameters and train the components responsible for processing the multimodal embeddings, described in \cref{eq:1}. More details are provided in \cref{app:addhyperparams}. For ViLBERT and UNITER we finetune the whole architecture at the same time.

All descriptions in \datasetacronym{} exceeding the maximum length of the three models are truncated. Due to their negligible amount, this does not affect performance significantly.

\section{Results}
\label{sec:results}

\begin{table}[t]
    \centering
    \begin{tabularx}{\columnwidth}{c | Y Y Y}
    \toprule
     & \textbf{all} & \textbf{video} & \textbf{static} \\
     \midrule
     \rowcolor{Gray}
      \multicolumn{4}{c}{\textsc{zero-shot}} \\
        CLIP & 22.4 & 15.6 & 47.8 \\
      \rowcolor{Gray}
      \multicolumn{4}{c}{\textsc{fine-tuning}} \\
        CLIP & 24.3 & 17.1 & 51.3 \\
      \hline
        {\small\contextbatch{}} & 28.4 & 20.0 & \textbf{60.0} \\
      \hline
        {\small\contextmodule{}} & 27.7 & 19.6 & 58.4 \\
      \hline
        {\small\temporalemb{}} & \textbf{29.9} & \textbf{22.0} & 59.8 \\
    \midrule
    \rowcolor{Gray}
    
    \multicolumn{4}{c}{\textsc{zero-shot}} \\
        UNITER & 19.8 & 13.6 & 42.9 \\
      \rowcolor{Gray}
      \multicolumn{4}{c}{\textsc{fine-tuning}} \\
        UNITER & 21.9 & 14.4 & 50.1 \textbf{} \\
      \hline
        {\small\contextbatch{}} & 24.8 & 17.4 & 52.8 \\
      \hline
        {\small\contextmodule{}} & 24.4 & 16.7 & \textbf{53.0} \\
      \hline
        {\small\temporalemb{}} & \textbf{25.7} & \textbf{19.1} & 50.5 \\
        \midrule
        \rowcolor{Gray}
      \multicolumn{4}{c}{\textsc{zero-shot}} \\
        ViLBERT & 19.3 & 13.5 & 40.8 \\
      \rowcolor{Gray}
      \multicolumn{4}{c}{\textsc{fine-tuning}} \\
        ViLBERT & 20.9 & 13.1 & \textbf{49.9} \\
      \hline
        {\small\contextbatch{}} & 20.9 & 15.0 & 42.7 \\
      \hline
        {\small\contextmodule{}} & 22.3 & 16.1 & 45.6 \\
      \hline
        {\small\temporalemb{}} & \textbf{24.5} & \textbf{18.0} & 49.3 \\
      \bottomrule
    \end{tabularx}
    \caption{Performance (test accuracy) on \datasetacronym{} across two training regimes (zero-shot and fine-tuning), three models (CLIP, UNITER, ViLBERT) and 4 model variants. We report separate figures for all the examples and two disjoint subsets: video frames and static pictures.}
    \label{tab:numtbl-test}
\end{table}

In \cref{tab:numtbl-test}, we report the performance of the models from \cref{sec:methods} for all the test examples in \datasetacronym{} as well as for the subsets containing only video frames or static pictures (see \cref{app:val-scores} for validation scores). Note that the random chance baseline has an accuracy of 10\%. In what follows, we compare the results across several dimensions.

\paragraph{Zero-shot vs.\ fine-tuning.}
In the zero-shot setting, we observe that CLIP representations are surprisingly superior to UNITER/ViLBERT even though CLIP has separate streams to encode an image and its description.
In the simplest fine-tuning setting (i.e., if negatives are randomly sampled independent of the image set), we find that overall there is only a small increase in performance compared to zero-shot inference. This demonstrates that in the regime where images in the same set do not appear in the same batch during training, models cannot extrapolate how to leverage the visual context at inference time.

\paragraph{Adding context.}
For the fine-tuning regime, we observe instead a different trend once the visual context of the other images in a set is provided during training (\contextbatch{}): CLIP and UNITER receive a significant boost in performance (i.e. +14.4\% for CLIP), which is particularly accentuated for static pictures. On the other hand, ViLBERT's performance remains the same. Stacking a special module for contextualizing multimodal representations on top of the encoders (\contextmodule{}), instead, yields gains for ViLBERT compared to \contextbatch{}, whereas CLIP and UNITER are unaffected (slight drop). This shows that all models can exploit visual context, but different strategies (contrastive training or dedicated modules) may be necessary.

Finally, all three models achieve the highest performance when fine-tuned with both visual and temporal context. Adding temporal positional embeddings on top of the contextual module (\temporalemb{}) yields an accuracy of 29.9 for CLIP, 25.7 for UNITER and 24.5 for ViLBERT. Crucially, even the best-performing models lag significantly behind the (micro-averaged) human accuracy of 90.8 (cf.\ \cref{tab:humans}).
Hence, despite some limited ability to integrate context, models are currently incapable of the fine-grained reasoning and pragmatic inferences needed to solve \datasetacronym{}.


\paragraph{Pre-trained model.}
Across all model variants and training regimes, CLIP consistently achieves higher accuracy than ViLBERT or UNITER. This implies that a larger amount of parameters, pre-training examples or the contrastive objective are more beneficial than ViLBERT's or UNITER's more expressive model architecture. Thus, these results violate the expectations that attention between vision and language would be more suitable to jointly encode highly nuanced visual details and descriptions \cite{miech_thinking_2021}. Additionally UNITER slightly outperforms ViLBERT as its single-stream architecture might enable richer cross-modal interactions.

\paragraph{Video frames vs.\ static pictures.}
The highest accuracy on the subset of the data with video frames (20.9) is far lower than that for static pictures (59.4). This confirms that videos represent the main challenge in \datasetacronym{}, both because of the higher similarity of images in a set and of the particular factors of variation that help differentiate among them (cf.\ \cref{ssec:visionstats} and examples in \cref{sec:app-phenomena}).
Additionally, model performance on video frames seems to increase more consistently as more context (both visual and temporal) is provided, whereas there is no clear trend in the case of static pictures.

\paragraph{Error Analysis.}
On a broad level, we have seen that video frames are much more challenging for models.
Next, to identify more fine-grained causes for the overall low performance of the vision-and-language models on \datasetacronym{}, we compute the Pearson's correlation between accuracy and a series of possible explanatory variables. In particular, we find a weak negative correlation with the number of tokens in the description ($r = -0.11$) and a weak positive correlation with the average pair-wise Euclidean distance between CLIP encodings of the images in a set ($r = 0.22$), which represents visual similarity.

\begin{figure}
    \centering
    \includegraphics[width=\columnwidth]{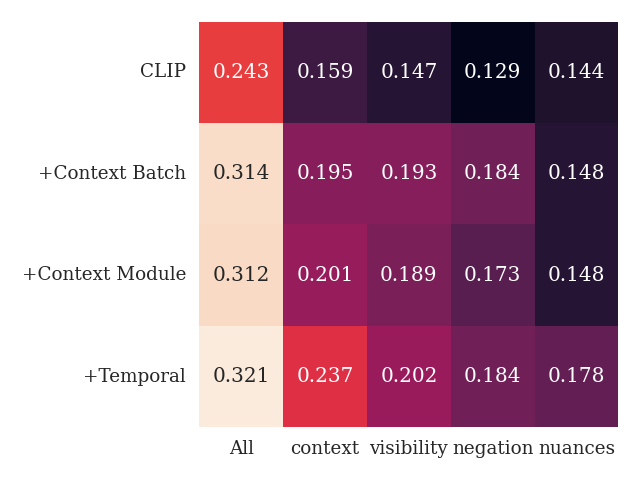}
    \caption{Performance of different CLIP variants (rows) on subsets of examples containing phenomena of interest (columns) in 1000 annotated validation examples. The hue of each cell indicates accuracy.}
    \label{fig:heatmap}
\end{figure}

By focusing on the 1000 annotated examples in \cref{tab:phenomena}
we observe a stark drop from overall performance on the subset of examples containing nuances, visibility/occlusion, and negation (\cref{fig:heatmap}).
This confirms insights from \newcite{kassner-schutze-2020-negated} and \newcite{hosseini-etal-2021-understanding} on the difficulty of modeling negation in text-only models.




\section{Conclusions and Future Work}
We created a new challenge, \datasetname{} (\datasetacronym{}), which is designed to evaluate the ability of vision-and-language models to integrate visual, pragmatic, and temporal context into their predictions. In particular, given a complex and nuanced \textit{contextual} description, a model is required to retrieve the corresponding image from a set of highly similar candidates. 
We benchmarked state-of-the-art bi-encoder and cross-encoder models, such as CLIP and ViLBERT.
Moreover, we proposed new variants of these models that are more suitable to solve this task, by augmenting them with a module to attend on the other images in a set and temporal embeddings. 
We found that \datasetacronym{} is highly challenging for all variants: even the best model (28.9) lags behind human performance (90.8) dramatically. Images sourced from video frames display the largest gap in performance.
The most challenging phenomena in \datasetacronym{} include pragmatics, negation, fine-grained distinctions between images, and occlusion among others.

\section{Acknowledgements}
\datasetacronym{} wouldn't have been possible without the herculean effort of the Amazon Mechanical Turkers and their feedback on the interface.
We also thank Emanuelle Bugliarello for his help with \href{https://github.com/e-bug/volta}{\textsc{Volta}}, an excellent codebase for several vision and language models.
We thank the members of SR's research group for their feedback on the ideas presented here.
\datasetacronym{} is funded by the Mila-Samsung grant program.
We thank Microsoft for providing us Azure credits.
SR acknowledges the support of the NSERC Discovery Grant program and the Facebook CIFAR AI Chair program.

\section{Ethics and Limitations}
We distribute the descriptions in \datasetacronym{} under MIT and adopt the licenses of the video and image sources on which our image sets build on top.
We report details about crowdsourcing such as payment and selection criteria in \cref{ssec:crowddesc} and \cref{app:anncrit}.
For the tested model variants, we only train a single run for each hyperparameter setting due to long run times.

\bibliography{literature}
\bibliographystyle{acl_natbib}


\appendix
\clearpage



\section{Length Distribution of the Image Descriptions}

\begin{figure}[h]
    \centering
    \includegraphics[scale=0.36]{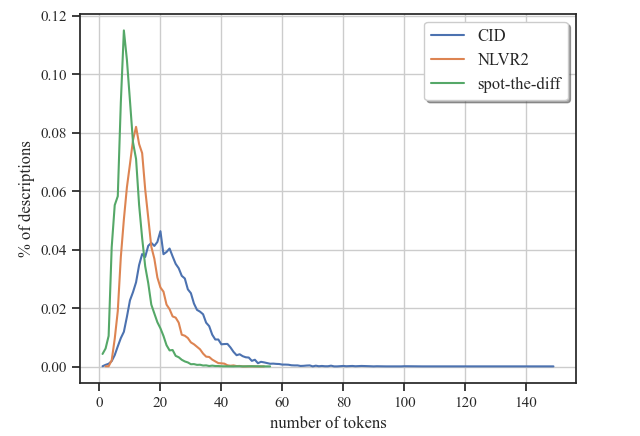}
    \caption{Distribution of the number of tokens across contextual descriptions in \datasetacronym{}.}
    \label{fig:numb_tokens}
\end{figure}

\section{Criteria for Selecting Annotators}
\label{app:anncrit}

We keep data quality high through entry requirements (English speaking country, over 98\% approval rate, etc.), qualification test, whitelisting workers and manually inspecting data.
Most importantly our two-stage setup also allowed us to automate monitoring data quality as we could measure the description and retrieval accuracy of workers and only whitelisted those with high accuracy.
We paid 0.25\$ per description and 0.1\$ per retrieval.

\section{Annotator Bias}
\label{app:bias}
The majority of descriptions in our test and validation split come from workers who did not work on the training set in order to avoid annotation bias.
Our validation set contains 502 descriptions from workers "seen" from the training set and 1,800 description from "unseen" workers.
In \cref{tab:bias} we can see that models perform slightly better on seen workers across our CLIP model variants.
\begin{table}[t]
    \centering
    \begin{tabularx}{\columnwidth}{c | Y Y}
    \toprule
     & \textbf{seen workers} & \textbf{unseen workers} \\
     \midrule
      \rowcolor{Gray}
      \multicolumn{3}{c}{\textsc{fine-tuning}} \\
        CLIP & \textbf{23.9} & 23.8 \\
      \hline
        {\small\contextbatch{}} & \textbf{34.5} & 29.0 \\
      \hline
        {\small\contextmodule{}} & \textbf{33.3} & 29.2 \\
      \hline
        {\small\temporalemb{}} & \textbf{32.1} & 30.8 \\
      \bottomrule
    \end{tabularx}
    \caption{Performance (accuracy) on two subsets of the distinct validation split: seen workers (workers who also produced description on the train split) and unseen workers (who only worked on the test and validation data).}
    \label{tab:bias}
\end{table}

\section{Crowdsourcing Interface}
\label{app:interface}
\begin{figure*}
    \centering
    \includegraphics[width=\textwidth,trim={0cm 0 0 2cm},clip]{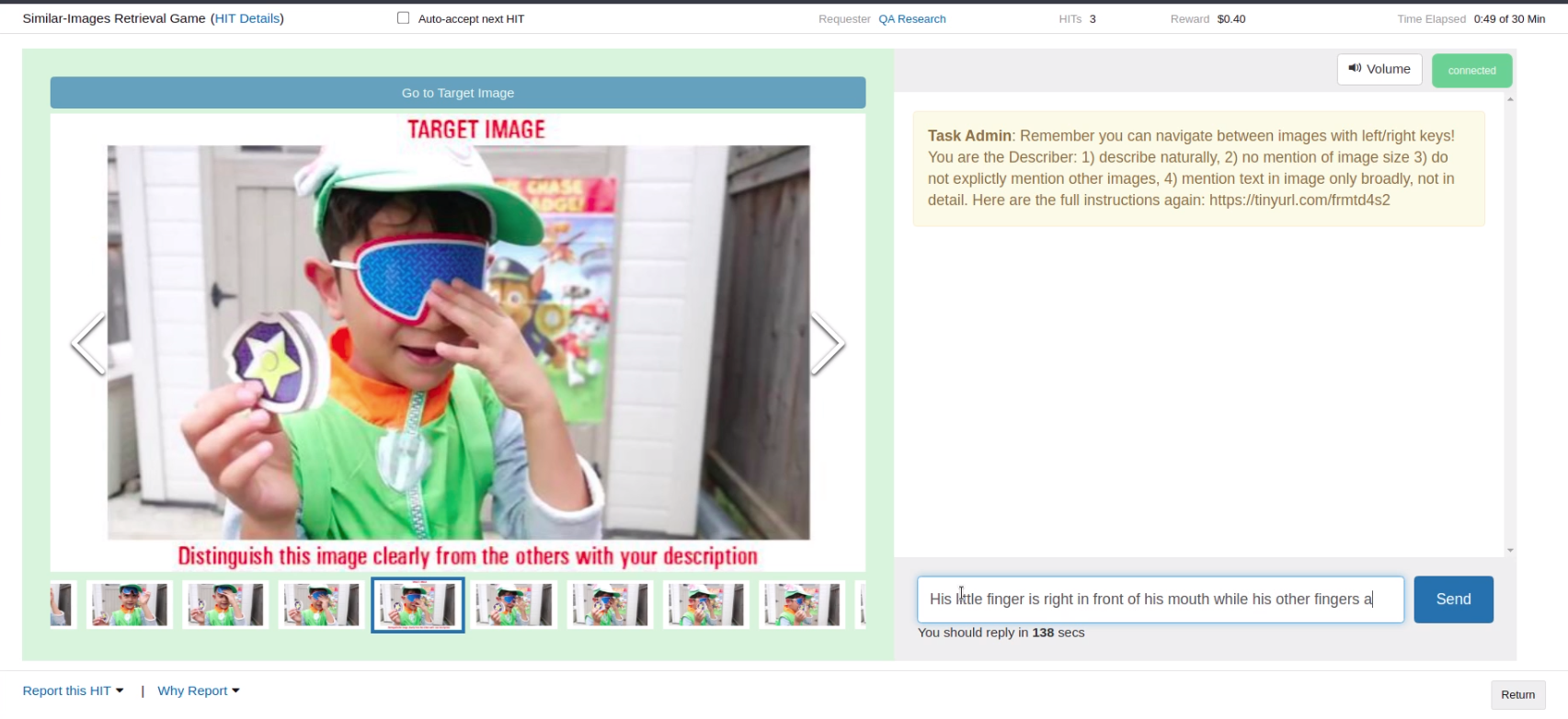}
    \caption{AMT interface for the describer task.}
    \label{fig:interface}
\end{figure*}

Our AMT interface for the description task can be seen in \cref{fig:interface}. The retriever interface looks conceptually similar, with a select-button for each image.
Note that workers see images almost in almost half of full-screen (opposed to the shown examples in this PDF) and can quickly go back and forth between consecutive frames with arrow-keys, making it significantly easier to spot and compare nuanced changes.

\section{Validation performance}
\label{app:val-scores}

\begin{table}[H]
    \centering
    \begin{tabularx}{\columnwidth}{c | Y Y Y}
    \toprule
     & \textbf{all} & \textbf{video} & \textbf{static} \\
     \midrule
     \rowcolor{Gray}
      \multicolumn{4}{c}{\textsc{zero-shot}} \\
        CLIP & 21.8 & 14.9 & 51.6 \\
      \rowcolor{Gray}
      \multicolumn{4}{c}{\textsc{fine-tuning}} \\
        CLIP & 23.4 & 17.3 & 50.2 \\
      \hline
        {\small\contextbatch{}} & 29.7 & 21.1 & 67.2 \\
      \hline
        {\small\contextmodule{}} & 29.9 & 21.4 & \textbf{67.2} \\
      \hline
        {\small\temporalemb{}} & \textbf{30.6} & \textbf{22.3} & 67.0 \\
    \midrule
    \rowcolor{Gray}
    \multicolumn{4}{c}{\textsc{zero-shot}} \\
        UNITER & 19.8 & 13.6 & 42.9 \\
      \rowcolor{Gray}
      \multicolumn{4}{c}{\textsc{fine-tuning}} \\
        UNITER & 23.8 & 17.5 & 51.2 \\
      \hline
        {\small\contextbatch{}} & 25.5 & 19.3 & 52.3 \\
      \hline
        {\small\contextmodule{}} & 24.8 & 18.9 & 50.7 \\
      \hline
        {\small\temporalemb{}} & \textbf{26.0} & \textbf{19.9} & \textbf{52.8} \\
    \midrule
    \rowcolor{Gray}
    \multicolumn{4}{c}{\textsc{zero-shot}} \\
       ViLBERT & 18.5 & 14.0 & 37.9 \\
      \rowcolor{Gray}
      \multicolumn{4}{c}{\textsc{fine-tuning}} \\
        ViLBERT & 21.9 & 16.1 & 46.7 \\
      \hline
        {\small\contextbatch{}} & 22.9 & 18.1 & 43.5 \\
      \hline
        {\small\contextmodule{}} & 23.5 & 18.9 & 43.5 \\
      \hline
        {\small\temporalemb{}} & \textbf{25.1} & \textbf{19.4} & \textbf{49.5} \\
      \bottomrule
    \end{tabularx}
    \caption{Performance (validation accuracy) on \datasetacronym{} across two training regimes (zero-shot and fine-tuning), three models (CLIP, UNITER, ViLBERT) and 4 model variants. We report separate figures for all the examples and two disjoint subsets: video frames and static pictures.}
    \label{tab:numtbl}
\end{table}

\section{Additional Hyper-parameters}
\label{app:addhyperparams}
The Transformer consists of 2 layers in CLIP variants and 4/5 layers in the ViLBERT/UNITER variants, both employing $\textrm{gelu}$ activation. The learning rate for the fine-tuning of the Transformer and linear heads is $2 \cdot 10^{-6}$ for the CLIP \contextmodule{}, $10^{-4}$ for CLIP \temporalemb{}, $2 \cdot 10^{-5}$ for both ViLBERT variants, and $6 \cdot 10^{-6}$ for both UNITER variants.
We use the Volta-framework \cite{bugliarello_multimodal_2021} for the standardized ViLBERT and UNITER model.

\section{Examples from \datasetacronym{} for all phenomena}
\label{sec:app-phenomena}
For each phenomenon we provide 1 example and a definition we used for annotation purposes.
Since most examples contain more than one phenomenon, some phenomena will be effectively showcased several times.
Note that we picked examples that are relatively easy to understand and spot differences in. 

\begin{figure}[h]
\centering
     \begin{subfigure}[t]{0.49\columnwidth}
         \centering
         \includegraphics[width=\textwidth]{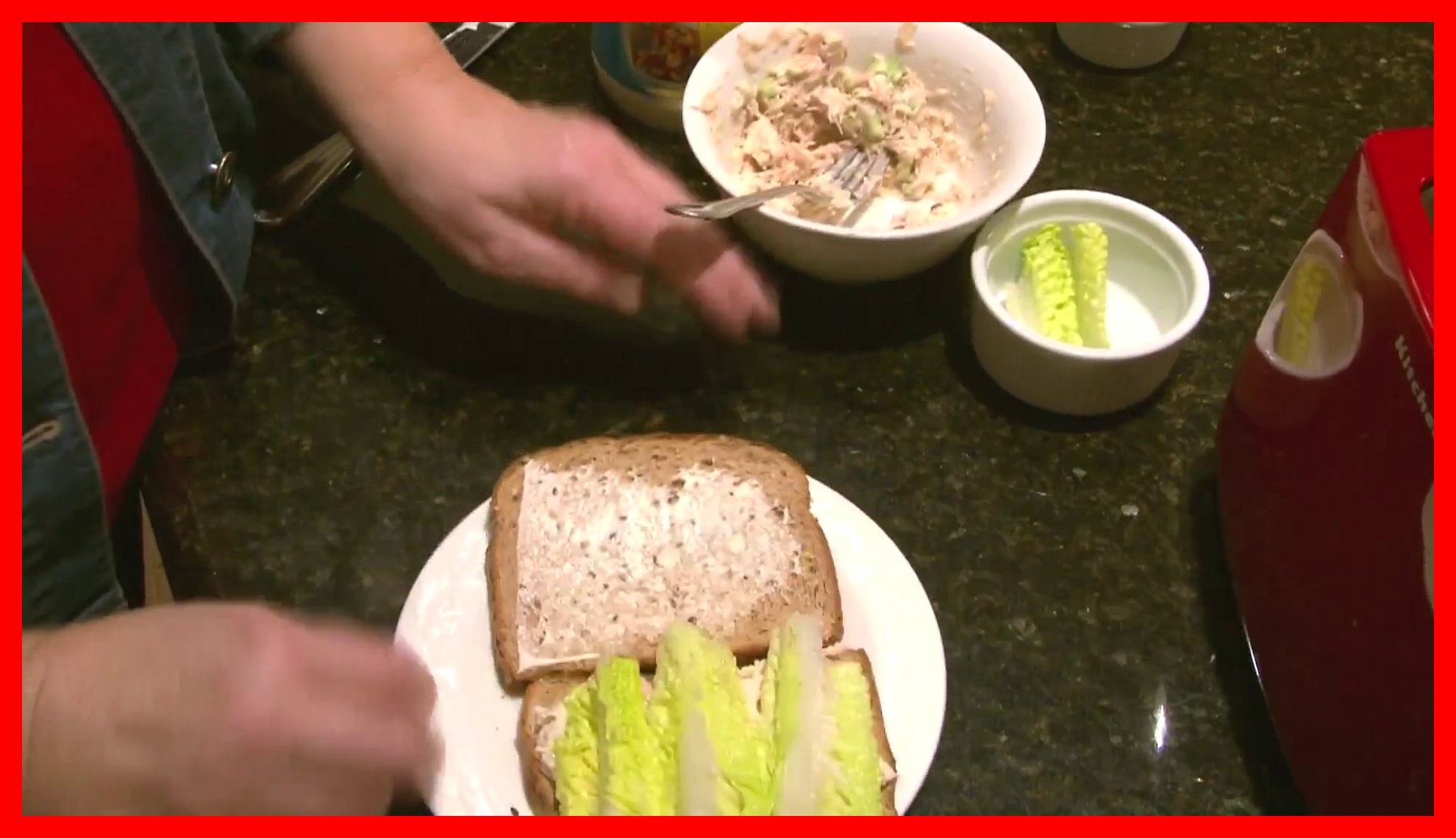}
         \caption{{Frame 6}}
     \end{subfigure}
     \hfill
     \begin{subfigure}[t]{0.49\columnwidth}
         \centering
         \includegraphics[width=\textwidth]{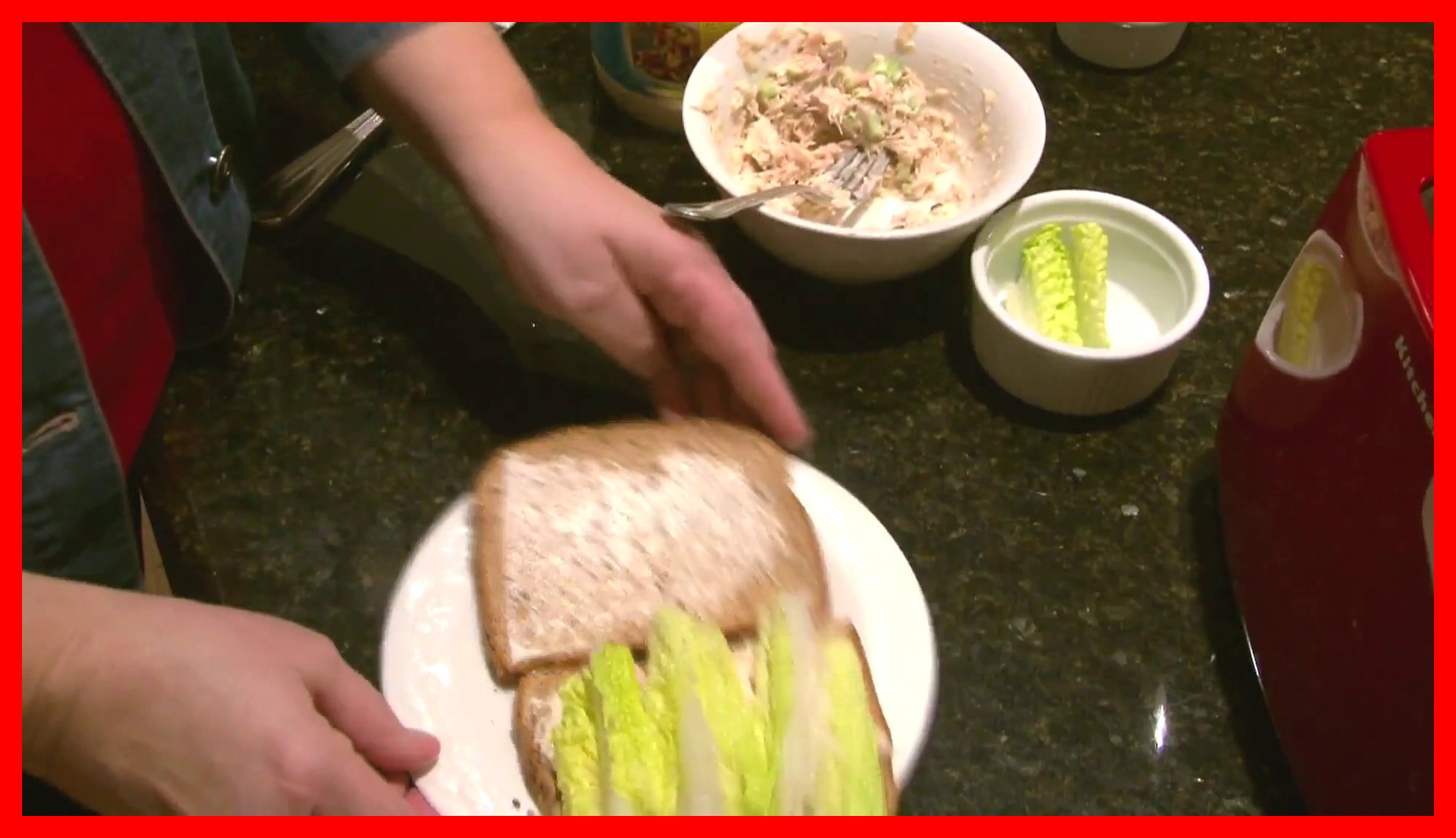}
         \caption{{Frame 7}}
     \end{subfigure}
     
     \begin{subfigure}[t]{0.49\columnwidth}
         \centering
         \includegraphics[width=\textwidth]{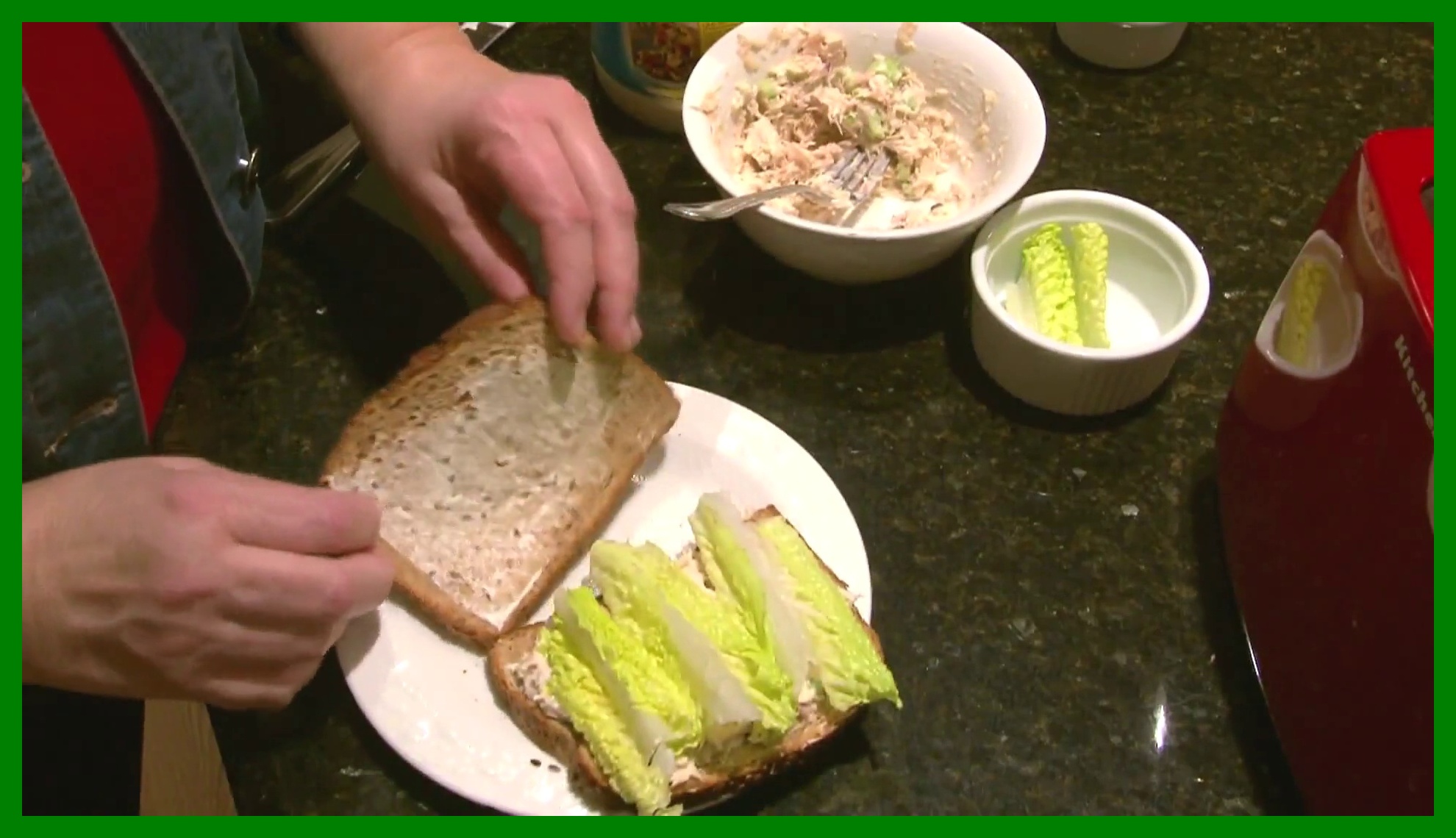}
         \caption{{Frame 8}}
     \end{subfigure}
     \hfill
     \begin{subfigure}[t]{0.49\columnwidth}
         \centering
         \includegraphics[width=\textwidth]{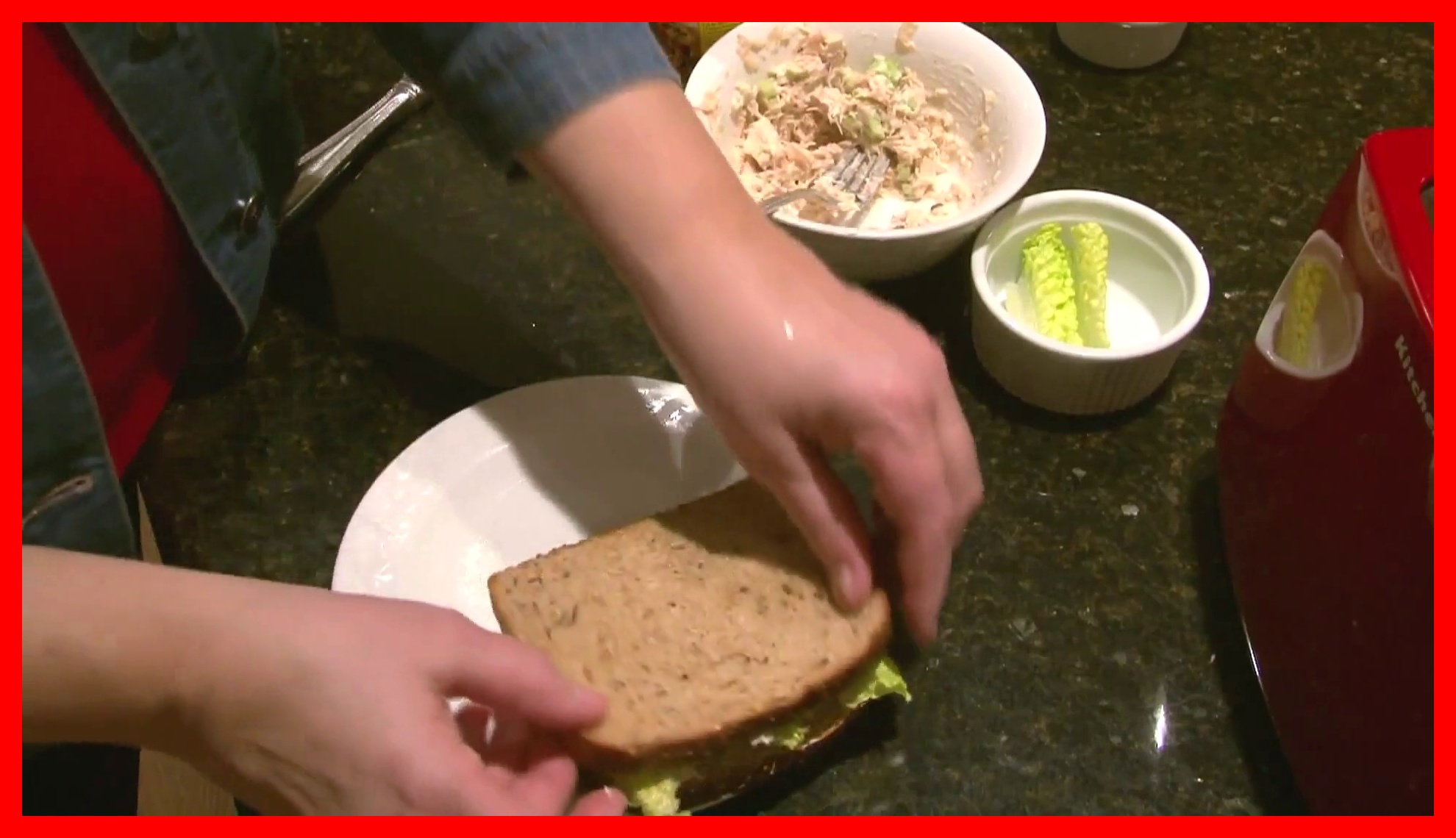}
         \caption{{Frame 9}}
     \end{subfigure}
\caption{
Example of \textbf{Context}:
\textit{``Both hands are on the piece of bread closest to the person.''}
Note: This is contextual since since without any context of other images, the description is also literally true for Frame 9. A model might even score it higher since the direct visual appearance is closer to typical bread.
Definition: To understand the description, a listener has to consider other images and/or the speakers intention of describing only one of the images. In line with Grice's maxim of quality, a description is contextual if it is literally true for several images but we know it was intended for only one image. A description is also contextual if an objects cannot clearly be identified in the target image directly but only through cross-referencing other images.
}
\label{bread}
\end{figure}

\begin{figure}[h]
\centering
     \begin{subfigure}[t]{0.49\columnwidth}
         \centering
         \includegraphics[width=\textwidth]{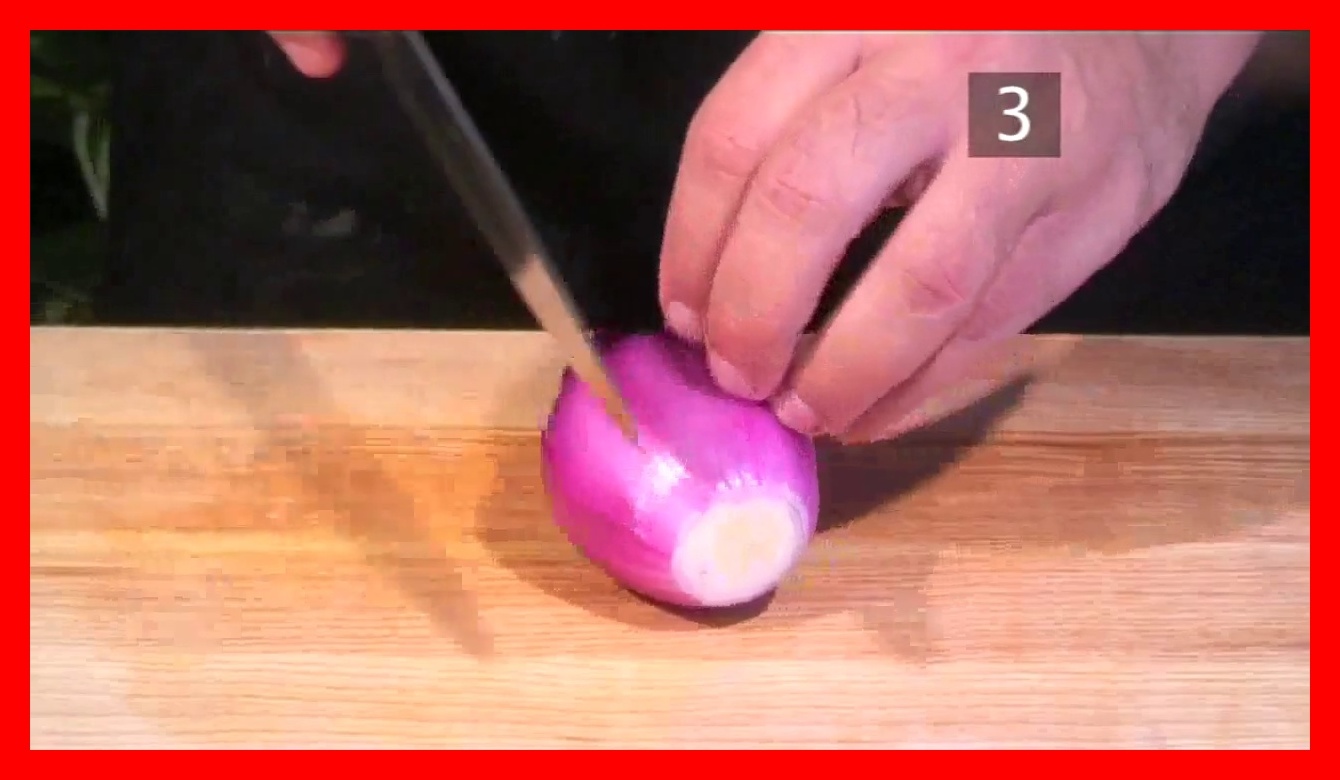}
         \caption{{Frame 3}}
     \end{subfigure}
     \hfill
     \begin{subfigure}[t]{0.49\columnwidth}
         \centering
         \includegraphics[width=\textwidth]{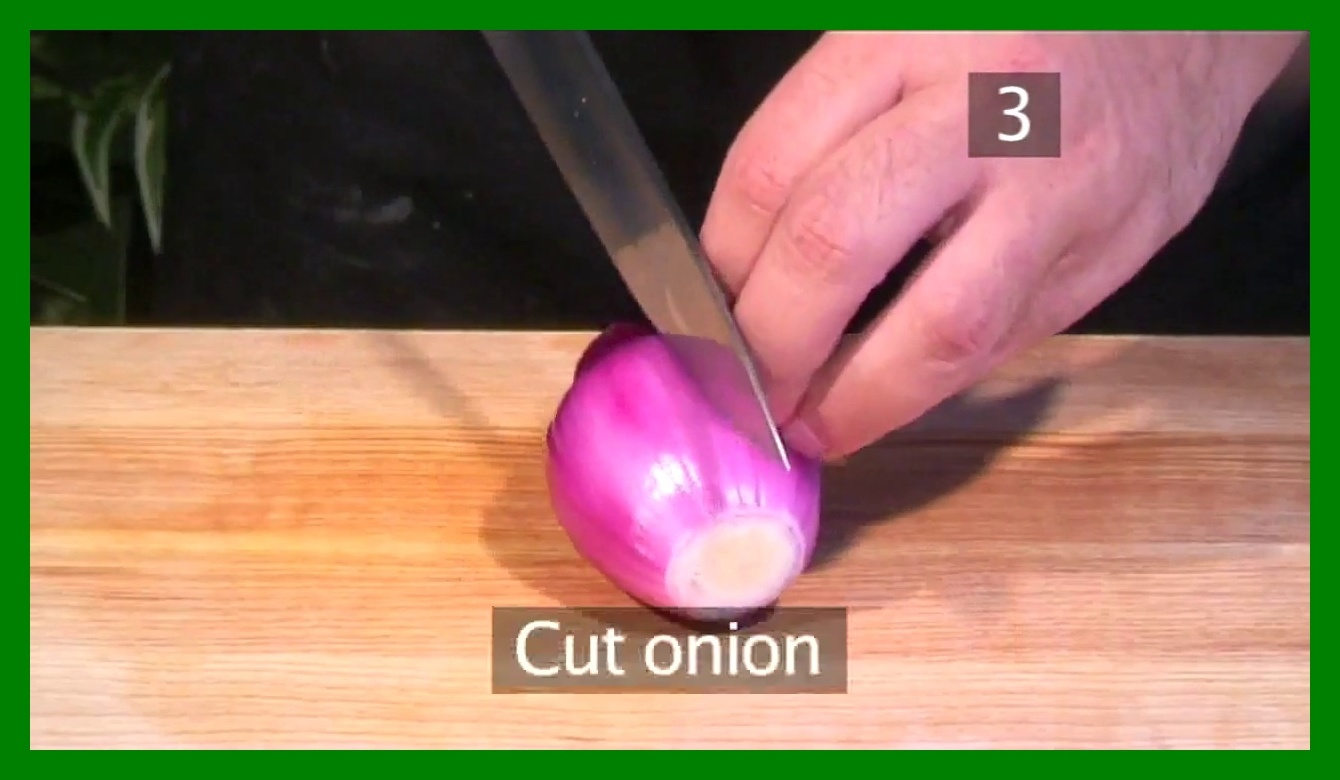}
         \caption{{Frame 4}}
     \end{subfigure}
     
     \begin{subfigure}[t]{0.49\columnwidth}
         \centering
         \includegraphics[width=\textwidth]{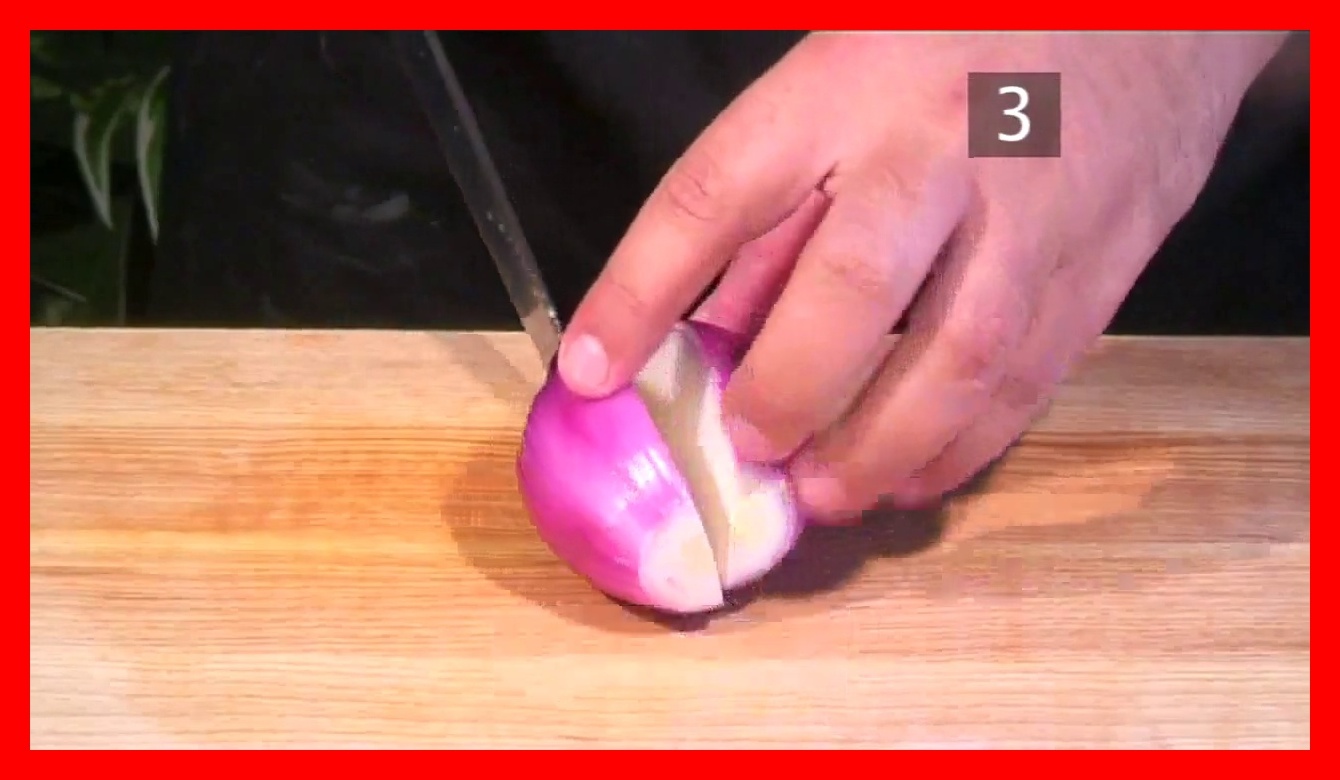}
         \caption{{Frame 5}}
     \end{subfigure}
     \hfill
     \begin{subfigure}[t]{0.49\columnwidth}
         \centering
         \includegraphics[width=\textwidth]{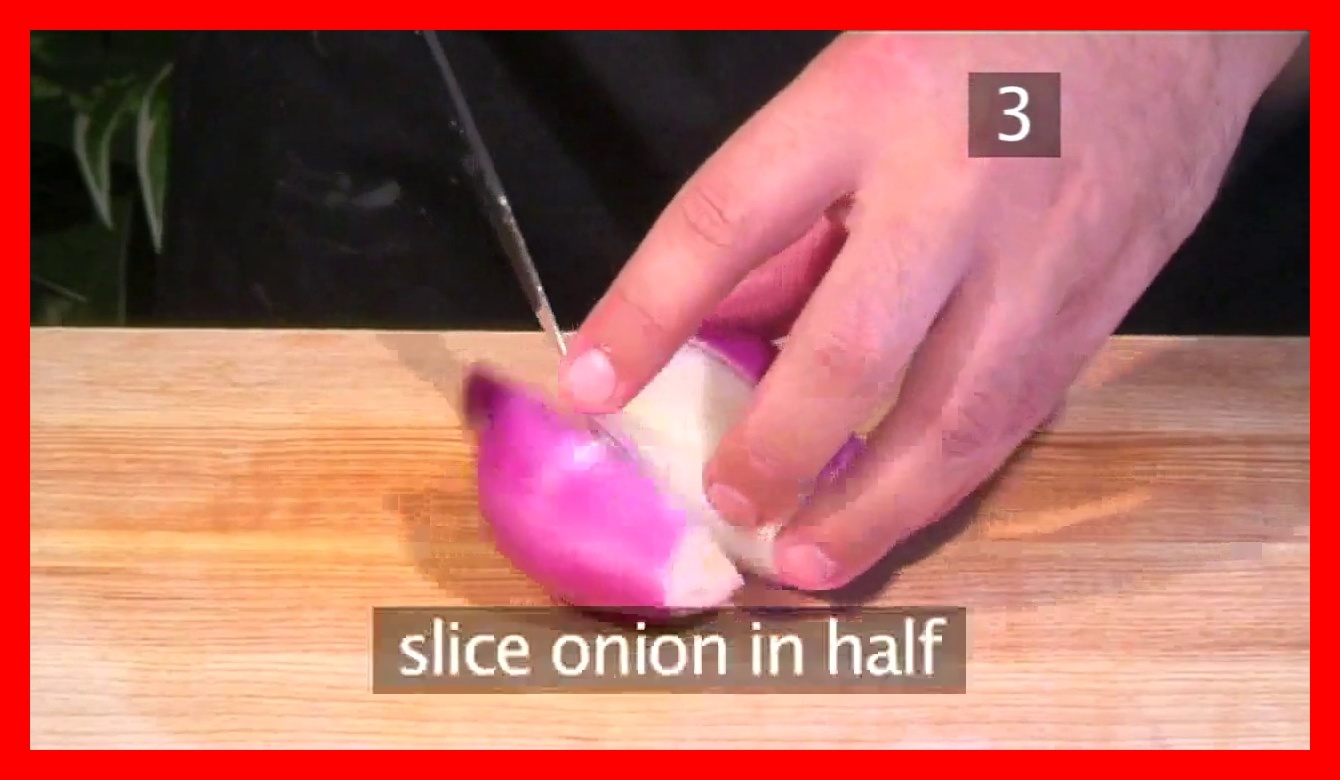}
         \caption{{Frame 6} }
     \end{subfigure}
\caption{
Example of \textbf{negation}:
\textit{``The knife is most centrally placed to insert into the onion \textbf{without} having fully cut deeply into it yet.''}
Definition: Explicit linguistic negation ("not", "unseen", "non-") or negation quantifiers ("no person").
}
\end{figure}

\begin{figure}[t]
\centering
     \begin{subfigure}[t]{0.49\columnwidth}
         \centering
         \includegraphics[width=0.57\textwidth]{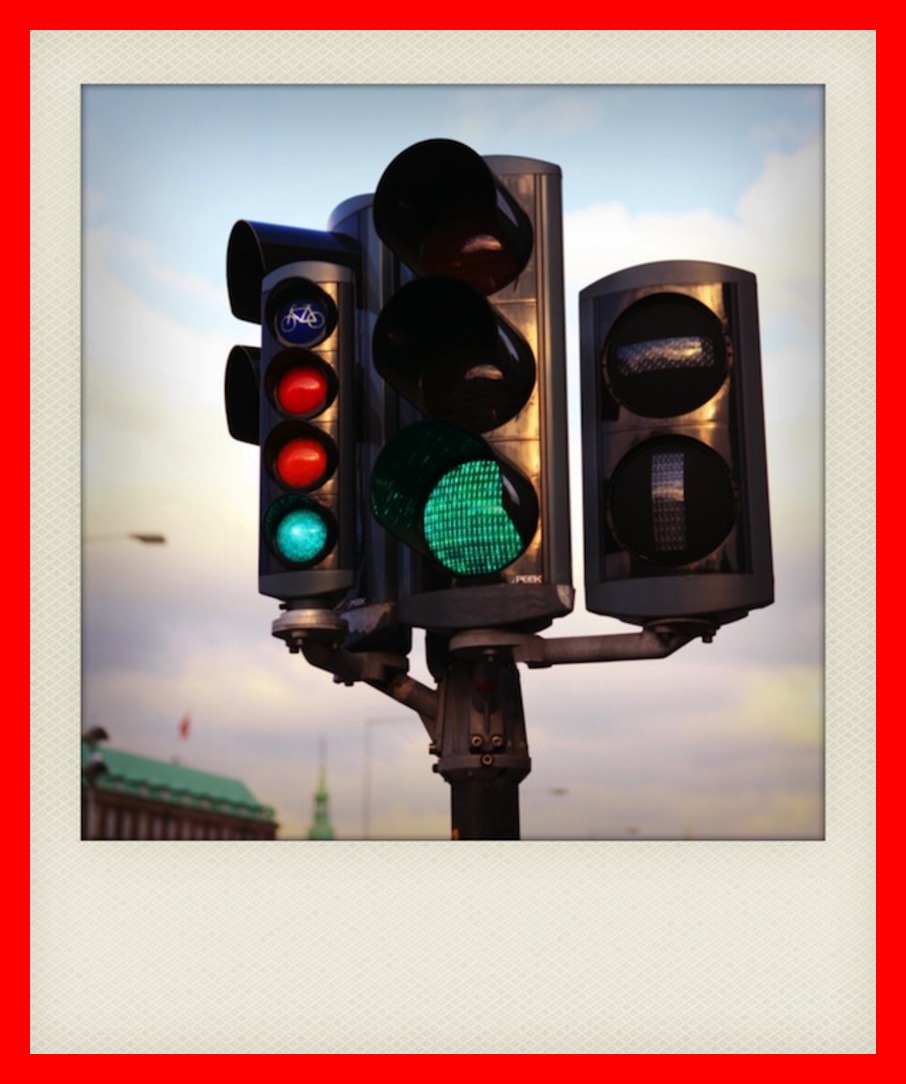}
         \caption{{Image 1}}
     \end{subfigure}
     \hfill
     \begin{subfigure}[t]{0.49\columnwidth}
         \centering
         \includegraphics[width=\textwidth]{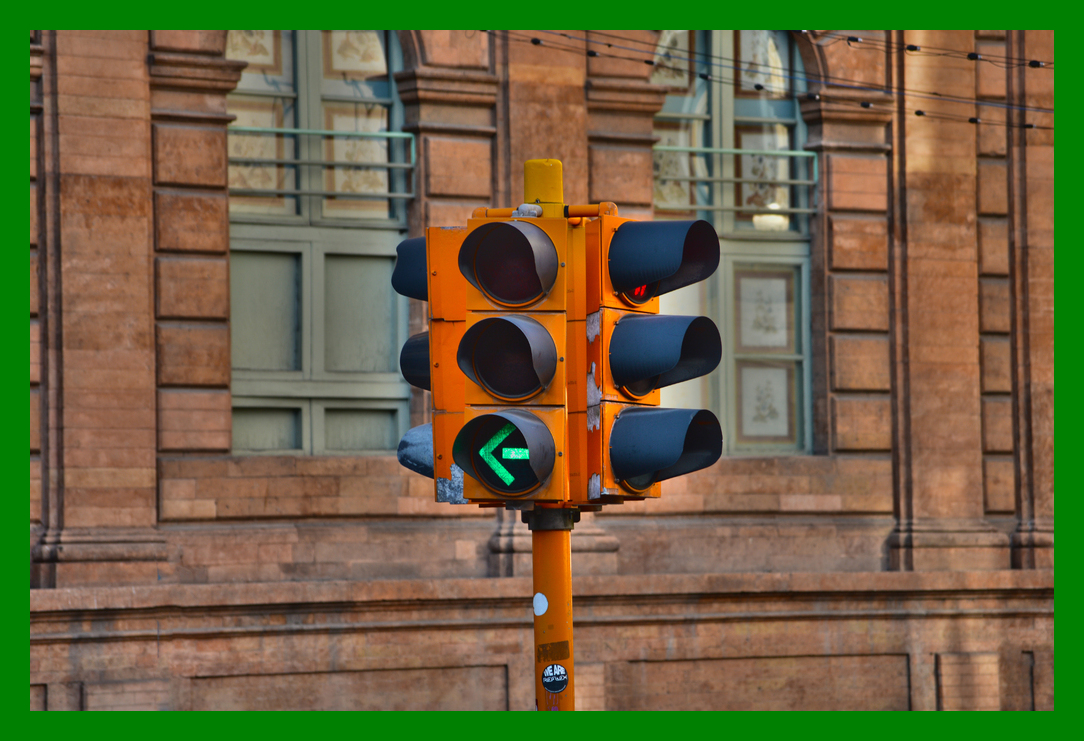}
         \caption{{Image 3}}
     \end{subfigure}
     
     \begin{subfigure}[t]{0.4\columnwidth}
         \centering
         \includegraphics[width=\textwidth]{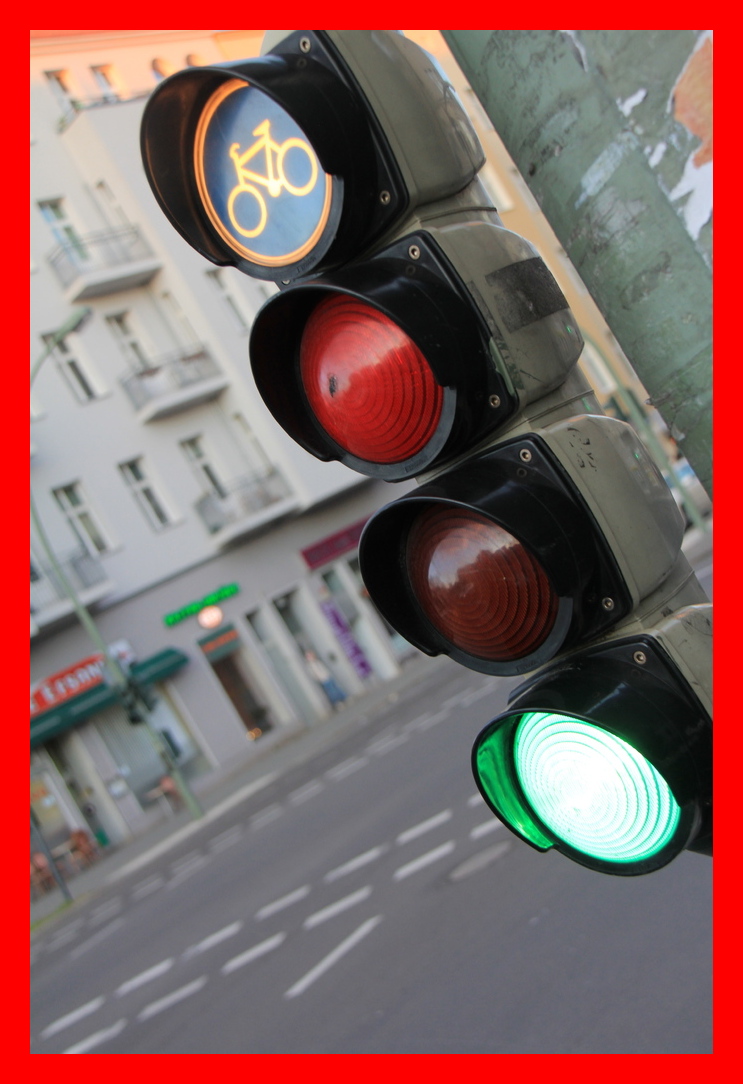}
         \caption{{Image 8}}
     \end{subfigure}
     \hfill
     \begin{subfigure}[t]{0.4\columnwidth}
         \centering
         \includegraphics[width=\textwidth]{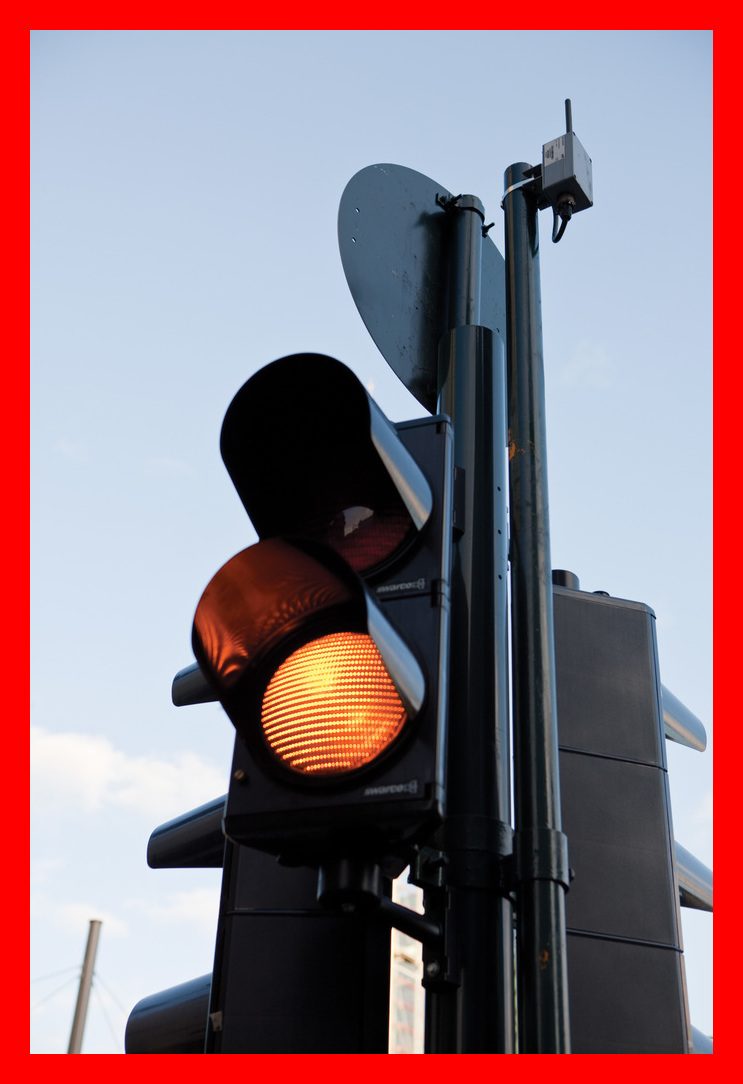}
         \caption{{Image 9}}
     \end{subfigure}
\caption{
Example of \textbf{quantifiers/quantities}:
\textit{``A yellow \textbf{3 way} traffic light with a green arrow on the side facing closest to the camera''}
Definition: We annotate for quantifiers (most, every, no, several,...) and absolute quantities ("five") as well as relative quantities (ratios like " a third of his hand").
}
\end{figure}

\begin{figure}[t]
\centering
     \begin{subfigure}[t]{0.49\columnwidth}
         \centering
         \includegraphics[width=\textwidth]{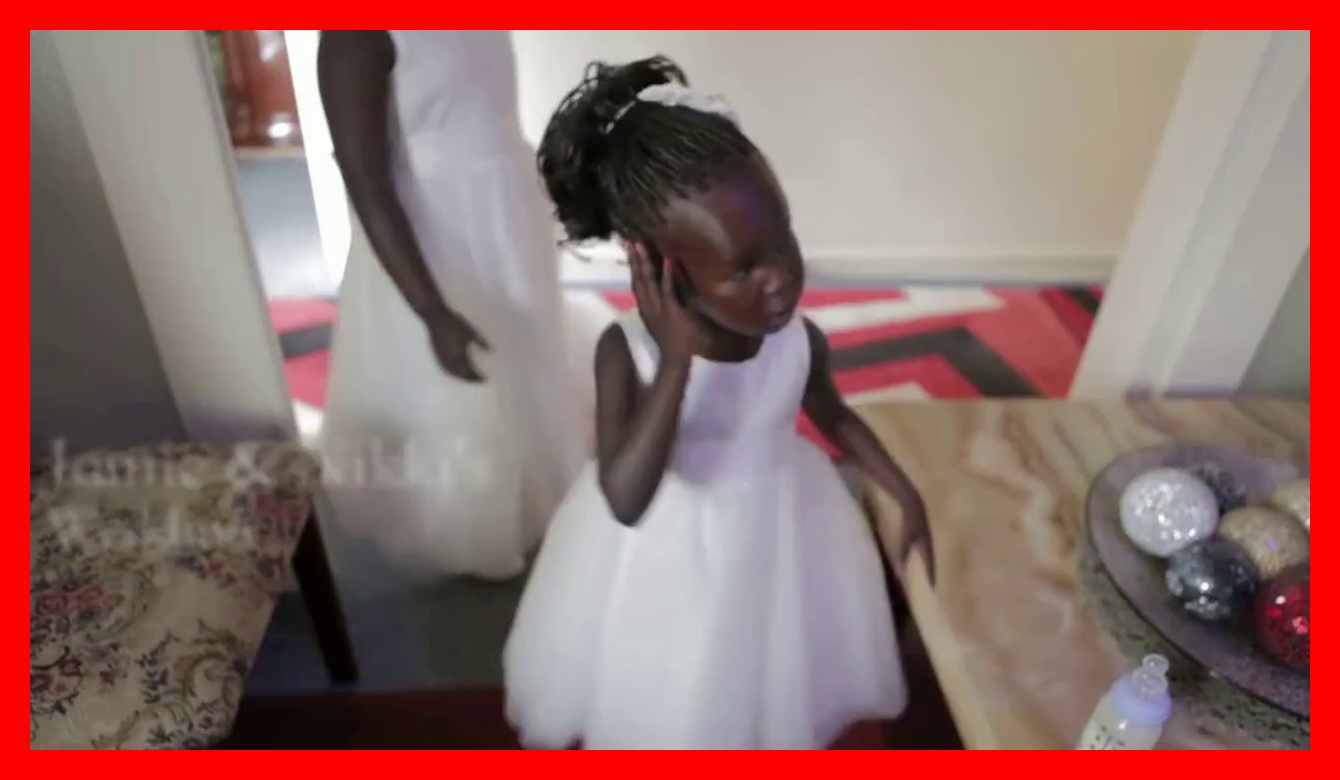}
         \caption{{Frame 4}}
     \end{subfigure}
     \hfill
     \begin{subfigure}[t]{0.49\columnwidth}
         \centering
         \includegraphics[width=\textwidth]{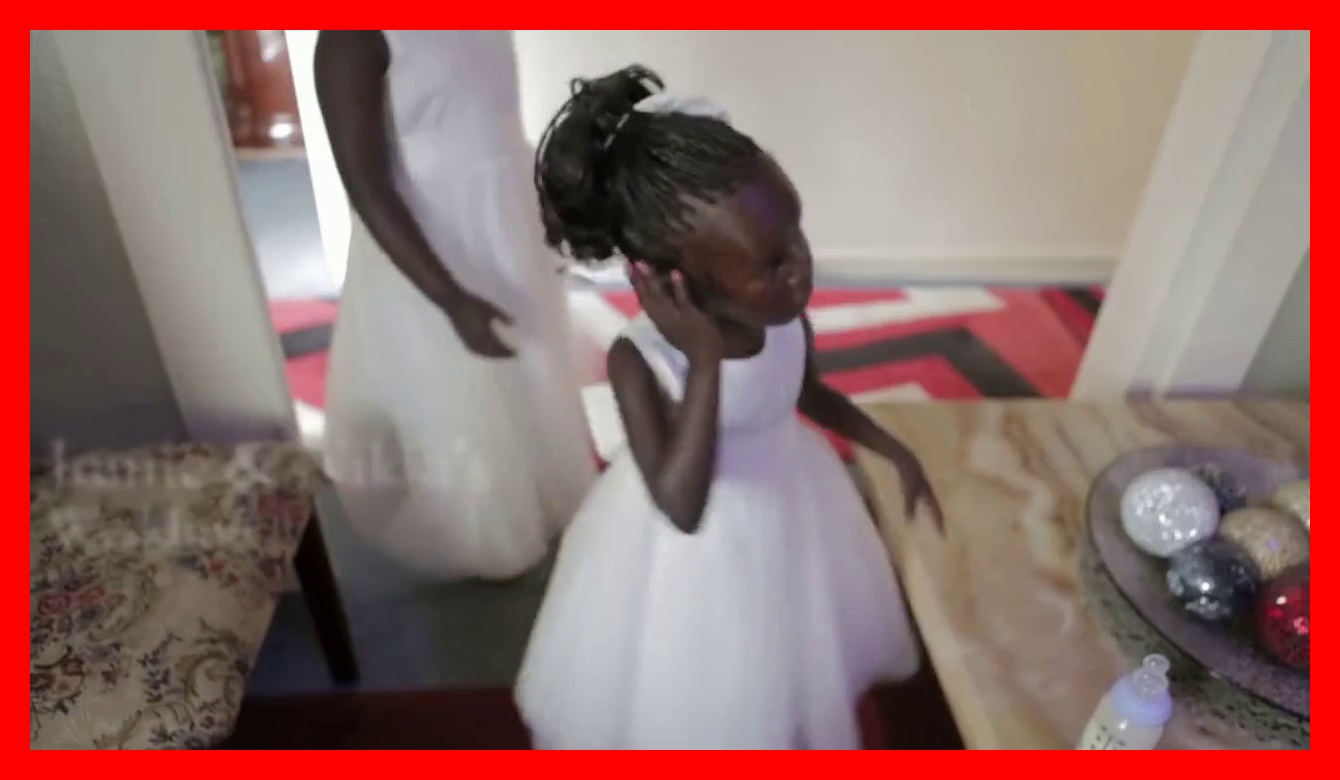}
         \caption{{Frame 5}}
     \end{subfigure}
     
     \begin{subfigure}[t]{0.49\columnwidth}
         \centering
         \includegraphics[width=\textwidth]{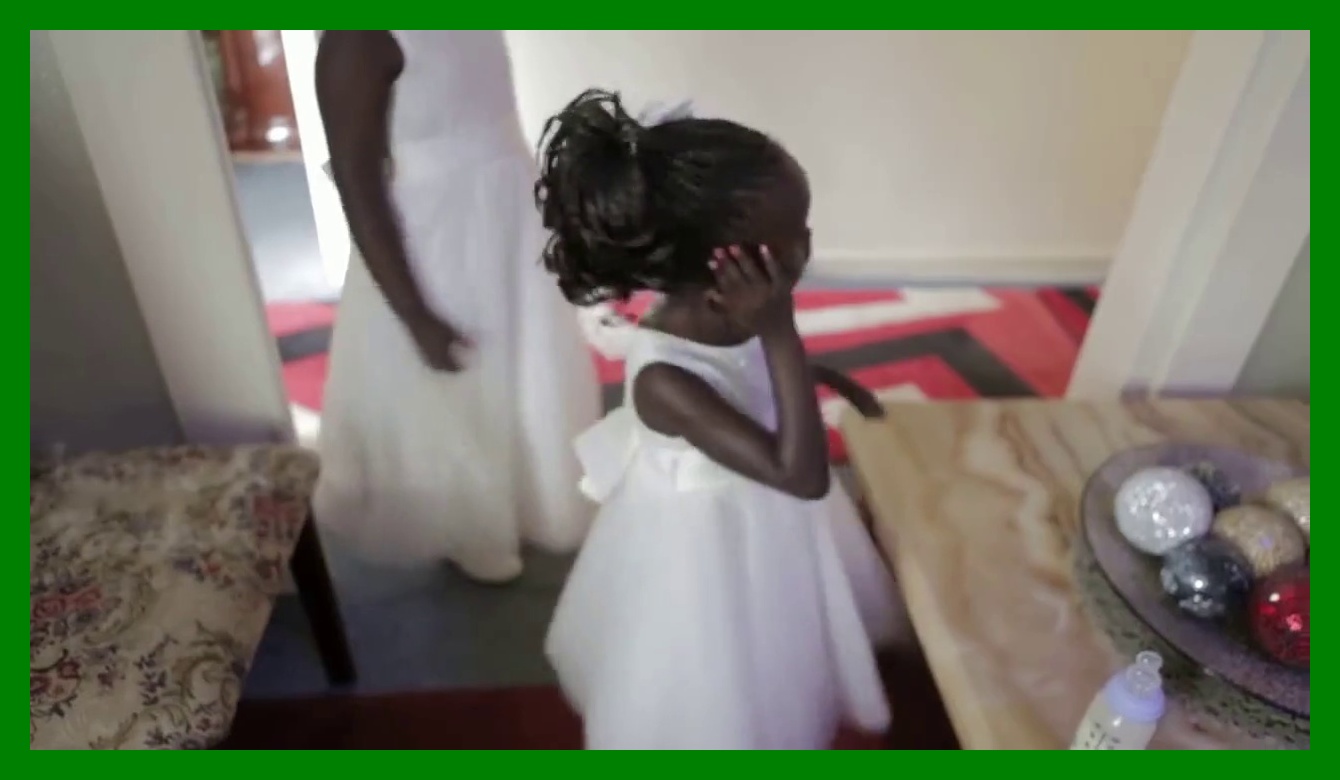}
         \caption{{Frame 6}}
     \end{subfigure}
     \hfill
     \begin{subfigure}[t]{0.49\columnwidth}
         \centering
         \includegraphics[width=\textwidth]{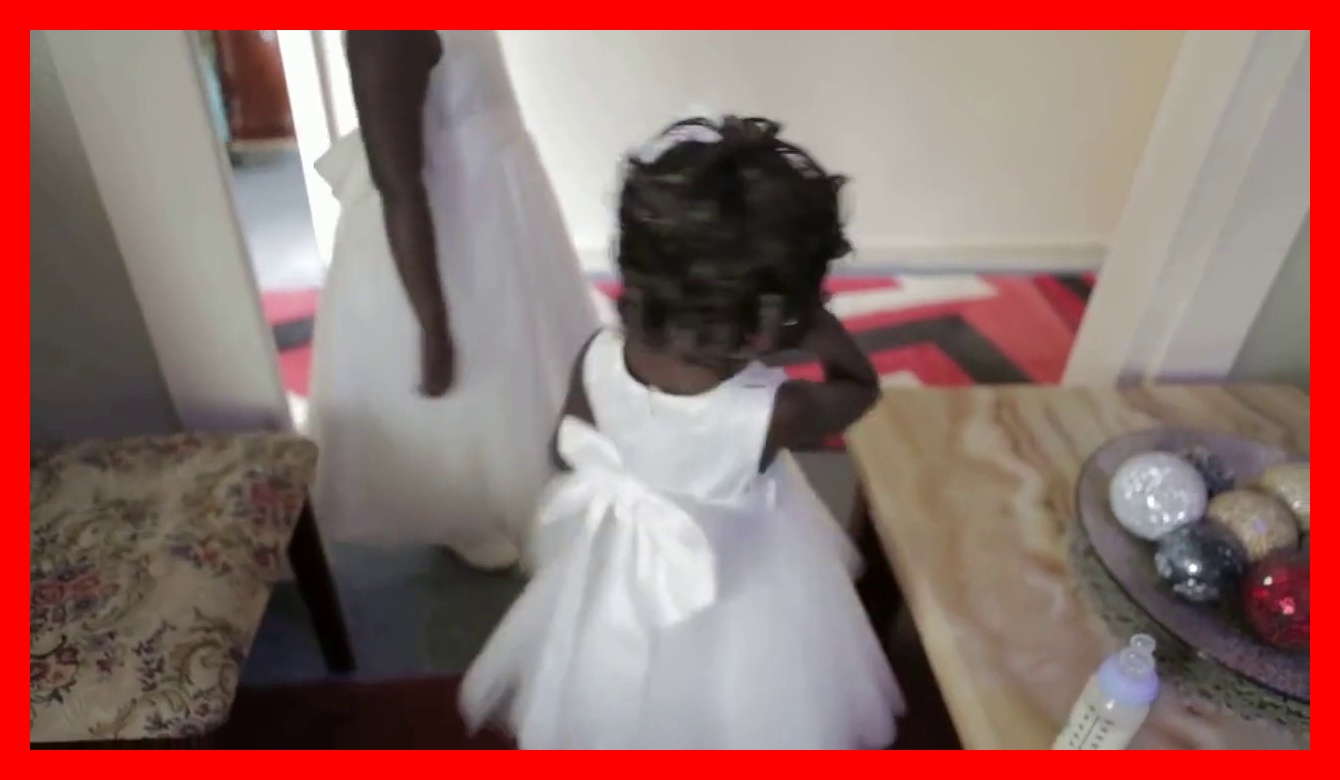}
         \caption{{Frame 7}}
     \end{subfigure}
\caption{
Example of \textbf{spatial relations/reasoning}:
\textit{``The small girl \textbf{in front} is looking \textbf{directly to the right} with her \textbf{right hand on the side} of her face.''}
Definition: Any relations or adjectives regarding space. Examples: "in the top left corner", "left to the chair", but also camera perspective, or body orientation ("turned towards...")
}
\end{figure}

\begin{figure}[t]
\centering
     \begin{subfigure}[t]{0.49\columnwidth}
         \centering
         \includegraphics[width=\textwidth]{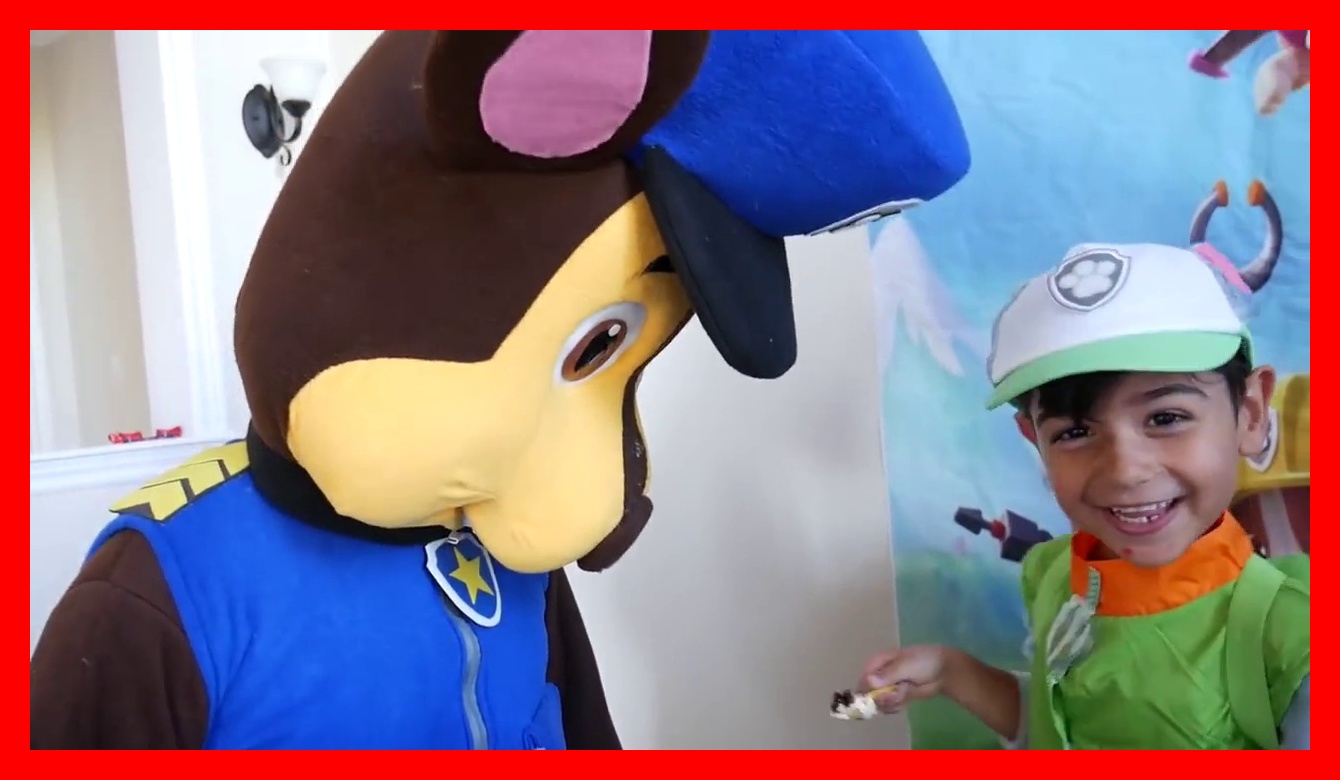}
         \caption{{Frame 6}}
     \end{subfigure}
     \hfill
     \begin{subfigure}[t]{0.49\columnwidth}
         \centering
         \includegraphics[width=\textwidth]{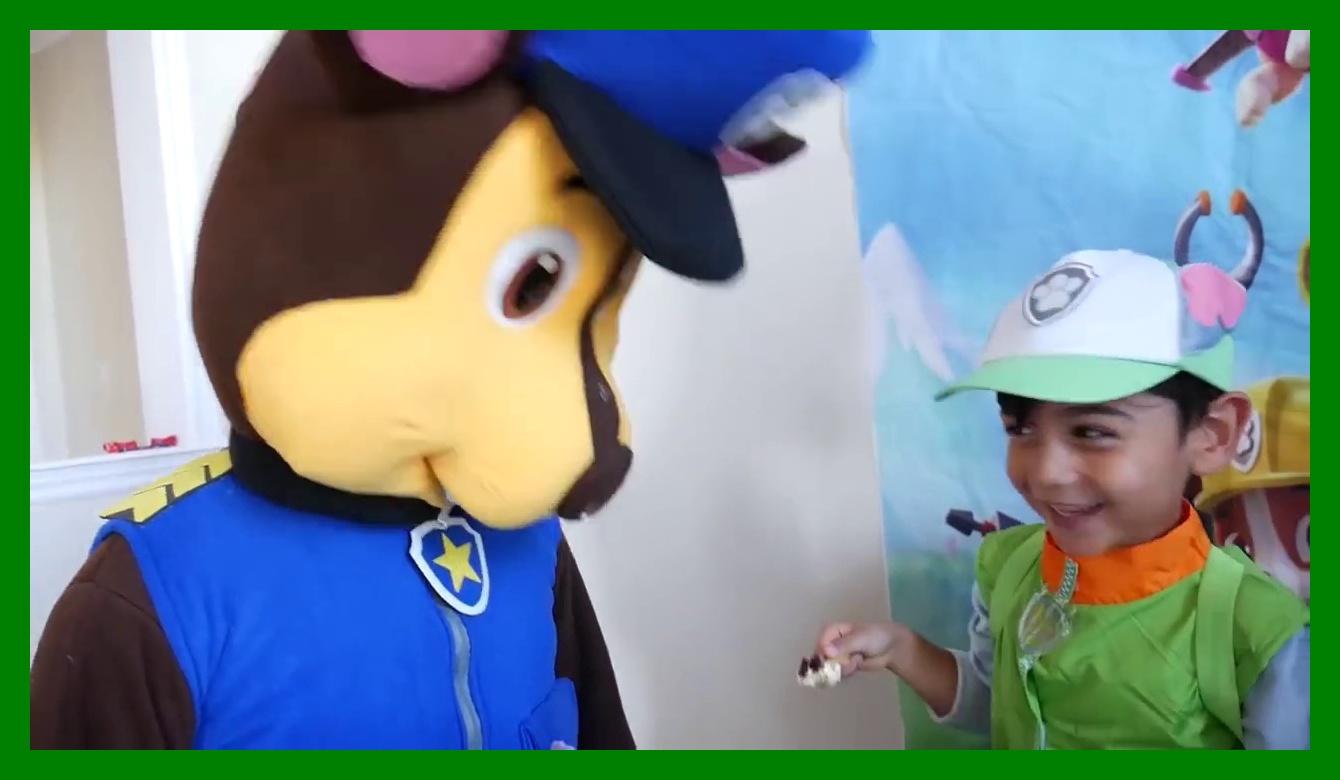}
         \caption{{Frame 7}}
     \end{subfigure}
     
     \begin{subfigure}[t]{0.49\columnwidth}
         \centering
         \includegraphics[width=\textwidth]{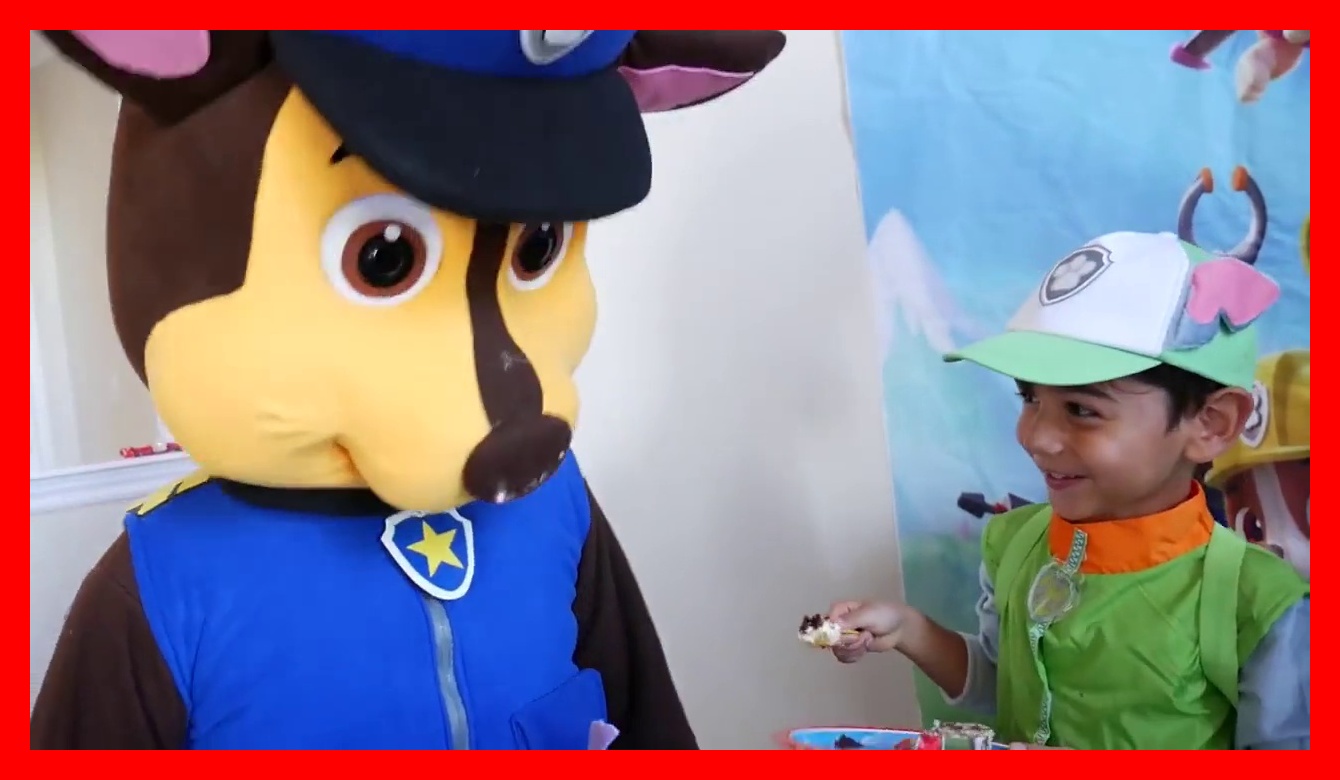}
         \caption{{Frame 8}}
     \end{subfigure}
     \hfill
     \begin{subfigure}[t]{0.49\columnwidth}
         \centering
         \includegraphics[width=\textwidth]{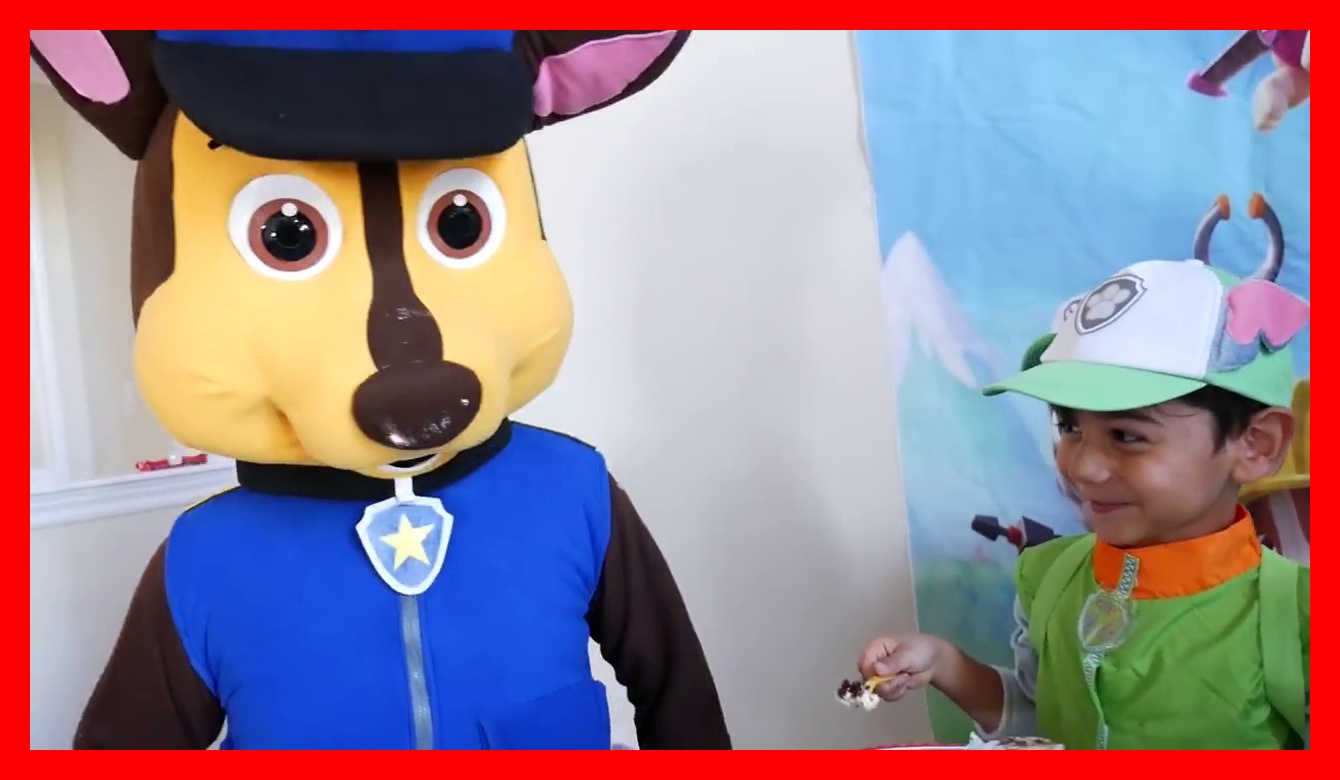}
         \caption{{Frame 9}}
     \end{subfigure}
\caption{
Example of \textbf{temporality}:
\textit{``
A smiling boy \textbf{just begins} to look towards the dog.''}
Definition: While most examples based on video frames implicitly require some temporal knowledge, we focus on explicit textual mentions of 1) temporal markers ("after", "during", "about to", etc) and 2) temporal verbs ("beginning to", "end to").
}
\end{figure}

\begin{figure}[t]
\centering
     \begin{subfigure}[t]{0.49\columnwidth}
         \centering
         \includegraphics[width=\textwidth]{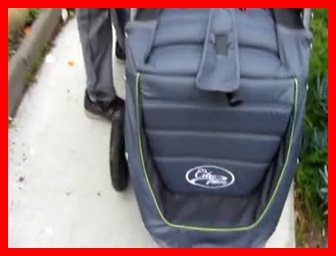}
         \caption{{Frame 7}}
     \end{subfigure}
     \hfill
     \begin{subfigure}[t]{0.49\columnwidth}
         \centering
         \includegraphics[width=\textwidth]{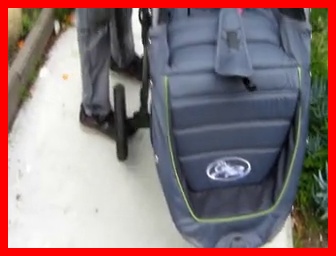}
         \caption{{Frame 8}}
     \end{subfigure}
     
     \begin{subfigure}[t]{0.49\columnwidth}
         \centering
         \includegraphics[width=\textwidth]{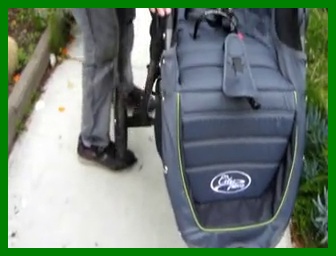}
         \caption{{Frame 9}}
     \end{subfigure}
     \hfill
     \begin{subfigure}[t]{0.49\columnwidth}
         \centering
         \includegraphics[width=\textwidth]{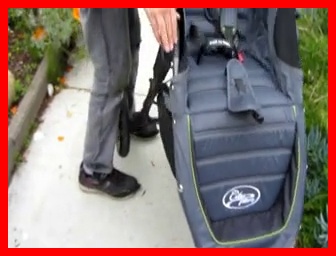}
         \caption{{Frame 10}}
     \end{subfigure}
\caption{
Example of \textbf{visibility/occlusion}:
\textit{``
The tire is directly \textbf{on top of} the person's right shoe and you can \textbf{just barely see} fingers at the top.
''}
Definition: A description that mentions objects/people being occluded, (partially) out of frame, or in the process of leaving the frame.
}
\end{figure}

\begin{figure}[t]
\centering
     \begin{subfigure}[t]{0.49\columnwidth}
         \centering
         \includegraphics[width=\textwidth]{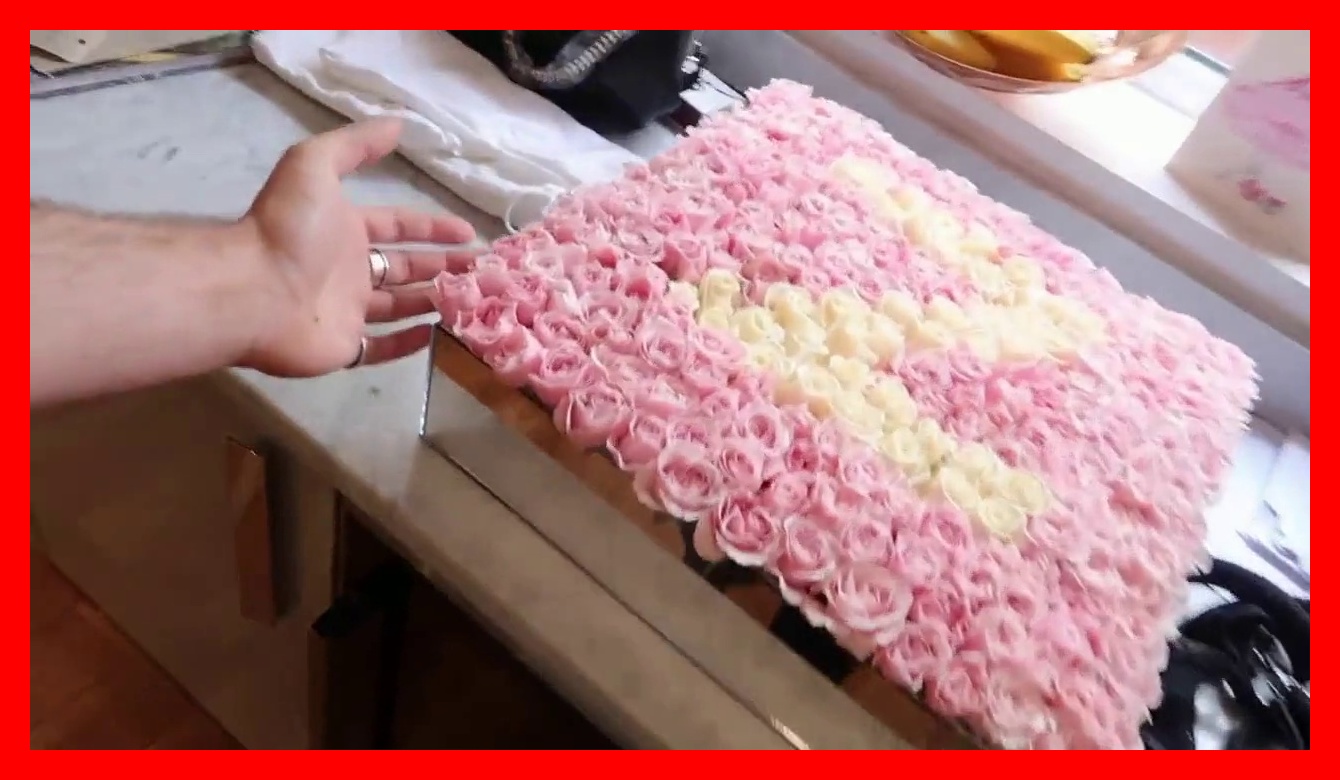}
         \caption{{Frame 4}}
     \end{subfigure}
     \hfill
     \begin{subfigure}[t]{0.49\columnwidth}
         \centering
         \includegraphics[width=\textwidth]{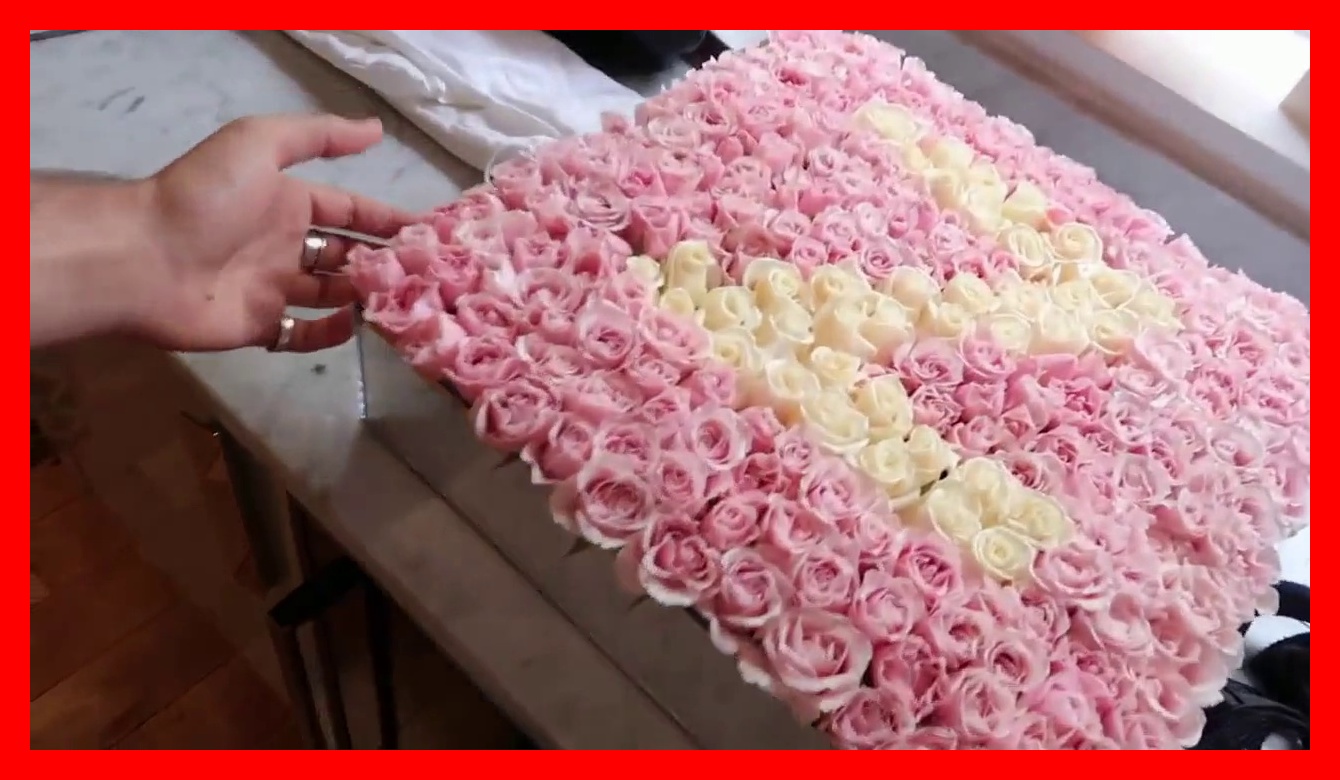}
         \caption{{Frame 5}}
     \end{subfigure}
     
     \begin{subfigure}[t]{0.49\columnwidth}
         \centering
         \includegraphics[width=\textwidth]{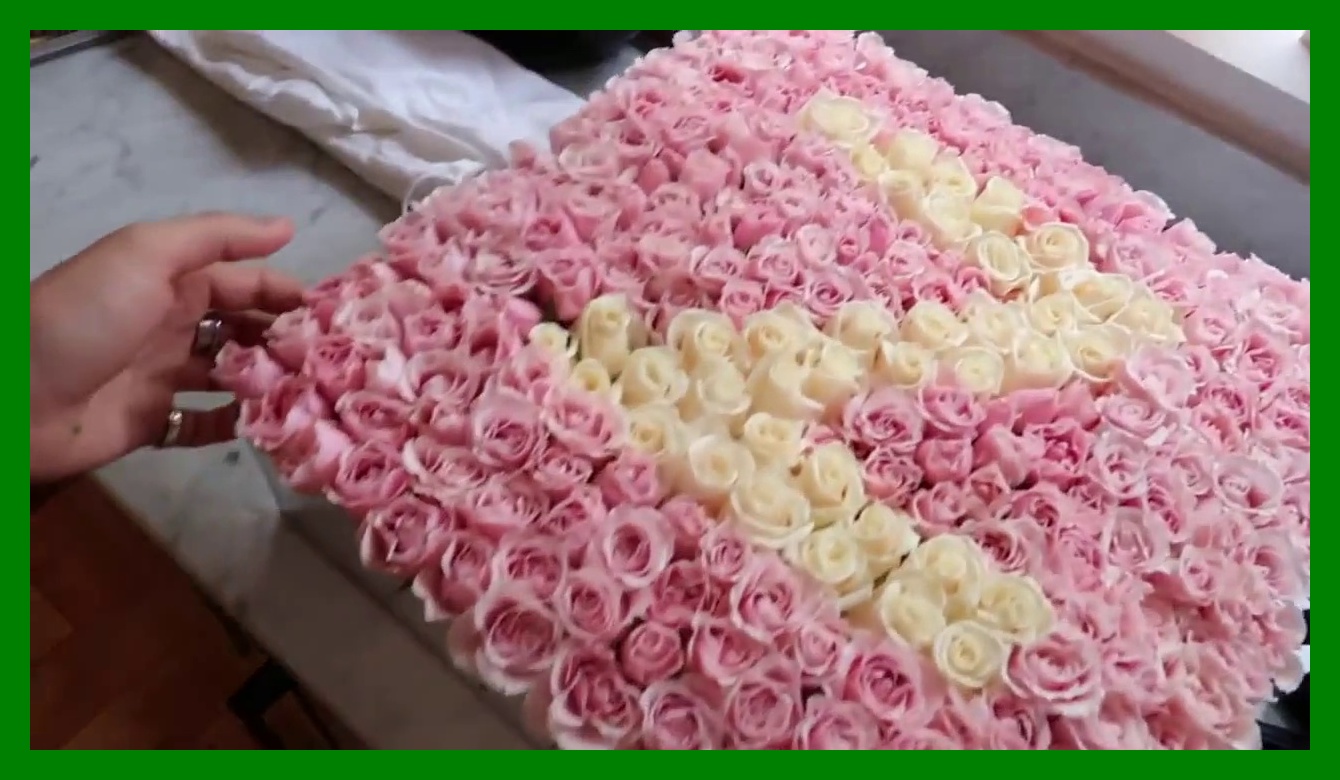}
         \caption{{Frame 6}}
     \end{subfigure}
     \hfill
     \begin{subfigure}[t]{0.49\columnwidth}
         \centering
         \includegraphics[width=\textwidth]{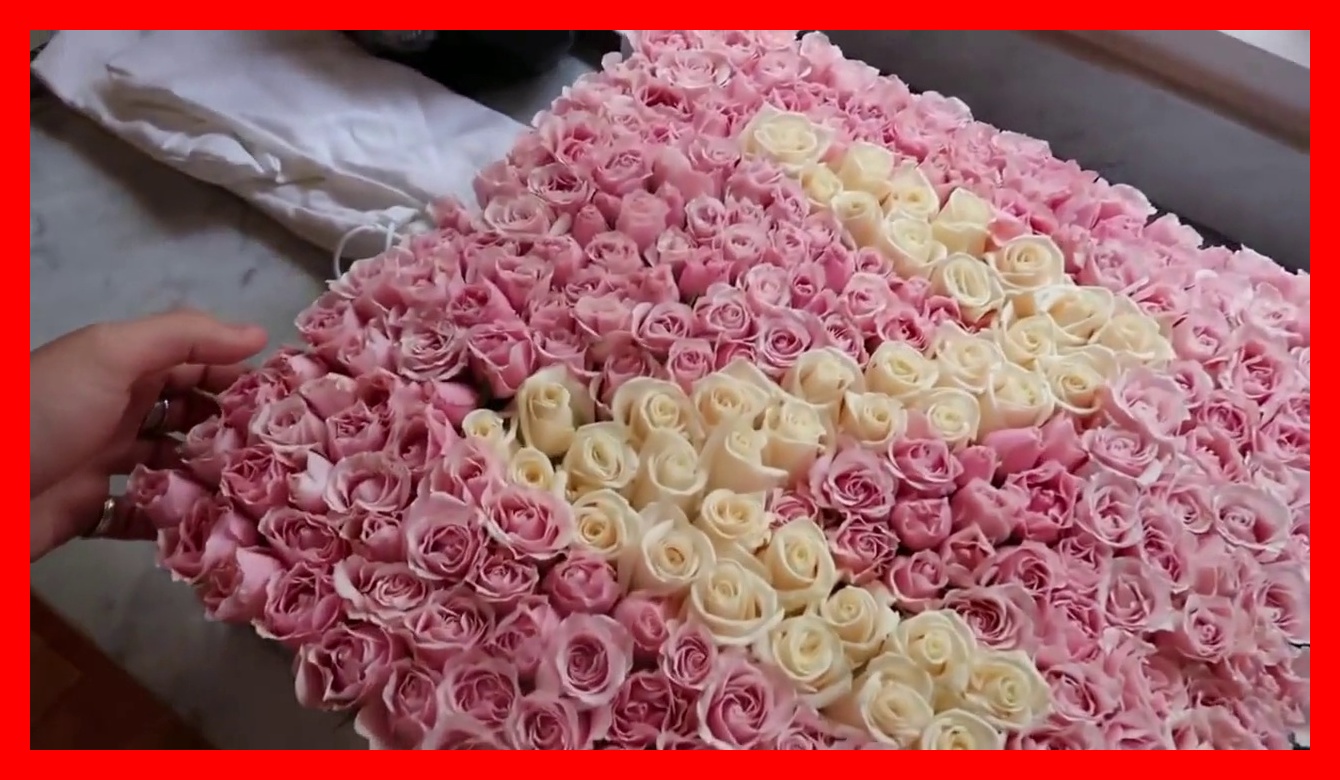}
         \caption{{Frame 7}}
     \end{subfigure}
\caption{
Example of \textbf{nuances} (we marked small details with red/green rectangles):
\textit{``
The person's palm is towards us and touching the left bottom corner of the cake. There is a \textbf{small amount of dark space between the right bottom corner of the photo and the edge of the cake}.
''}
Definition: Minor details, that are either a) not salient at all and would usually be left unmentioned and/or b) language reference is grounded on a small patch of pixels. Note that this phenomena is often linked with very minimally contrastive images.
}
\end{figure}

\begin{figure}[t]
\centering
     \begin{subfigure}[t]{0.49\columnwidth}
         \centering
         \includegraphics[width=\textwidth]{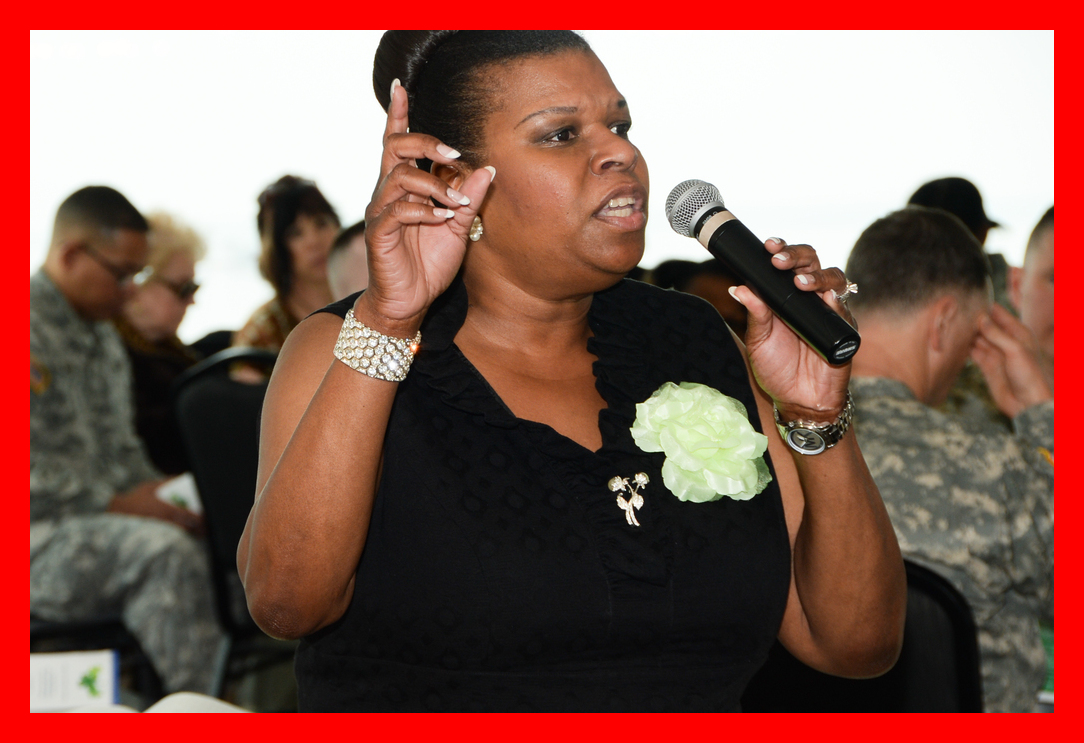}
         \caption{{Image 3}}
     \end{subfigure}
     \hfill
     \begin{subfigure}[t]{0.49\columnwidth}
         \centering
         \includegraphics[width=\textwidth]{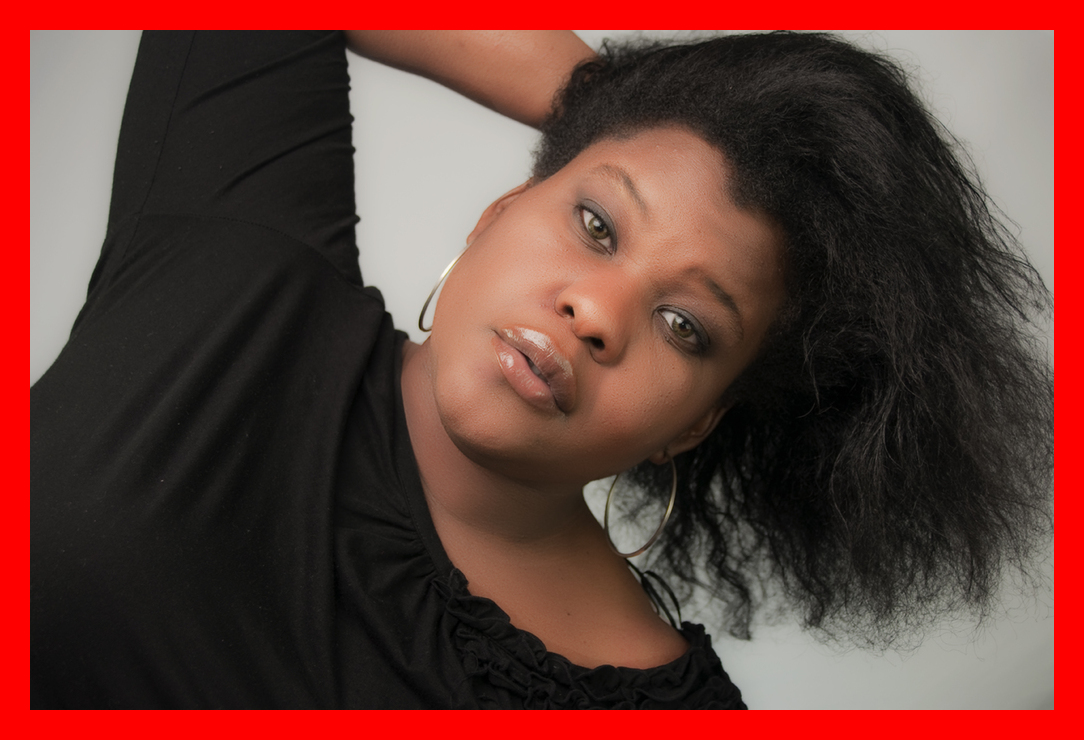}
         \caption{{Image 5}}
     \end{subfigure}
     
     \begin{subfigure}[t]{0.49\columnwidth}
         \centering
         \includegraphics[width=\textwidth]{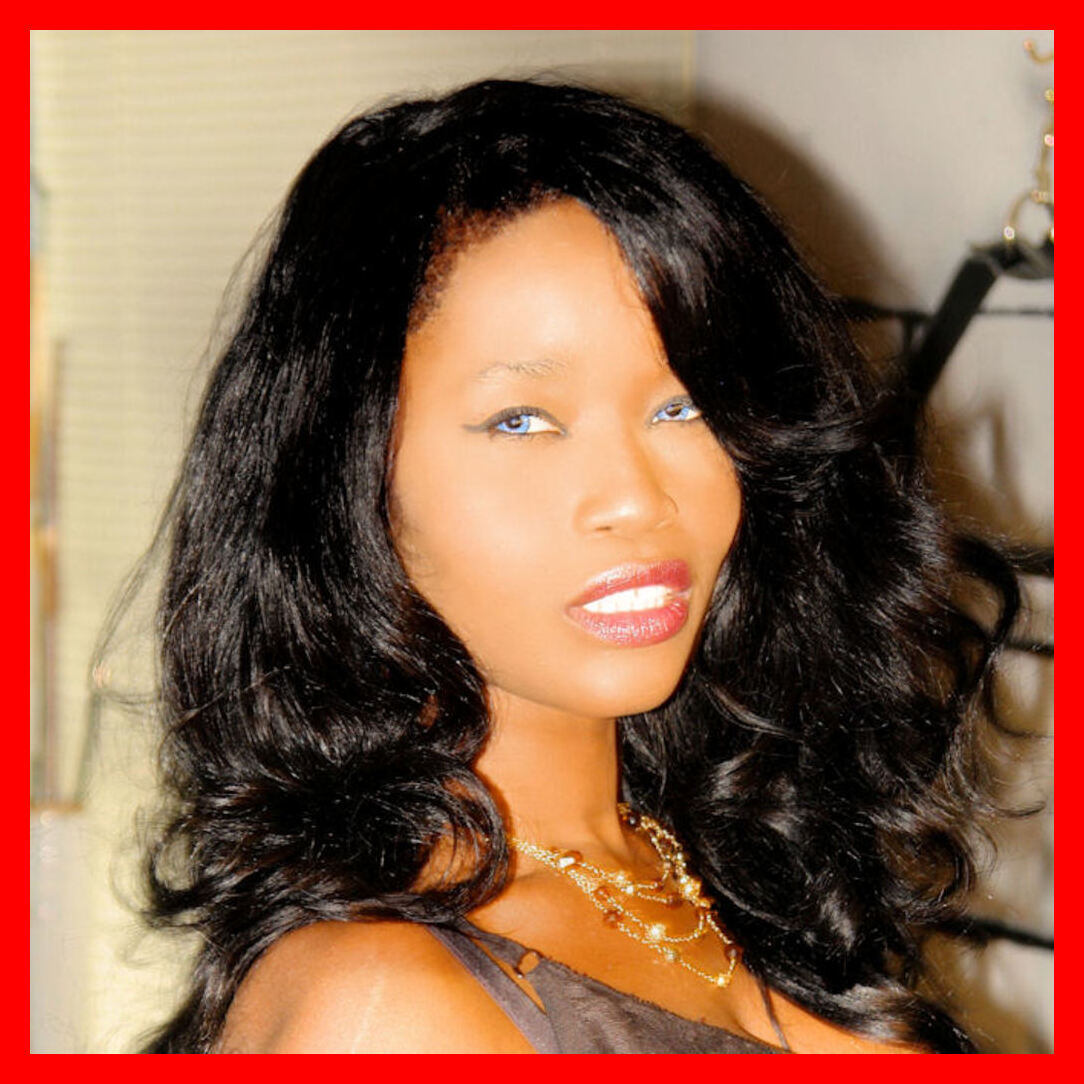}
         \caption{{Image 6}}
     \end{subfigure}
     \hfill
     \begin{subfigure}[t]{0.49\columnwidth}
         \centering
         \includegraphics[width=0.68\textwidth]{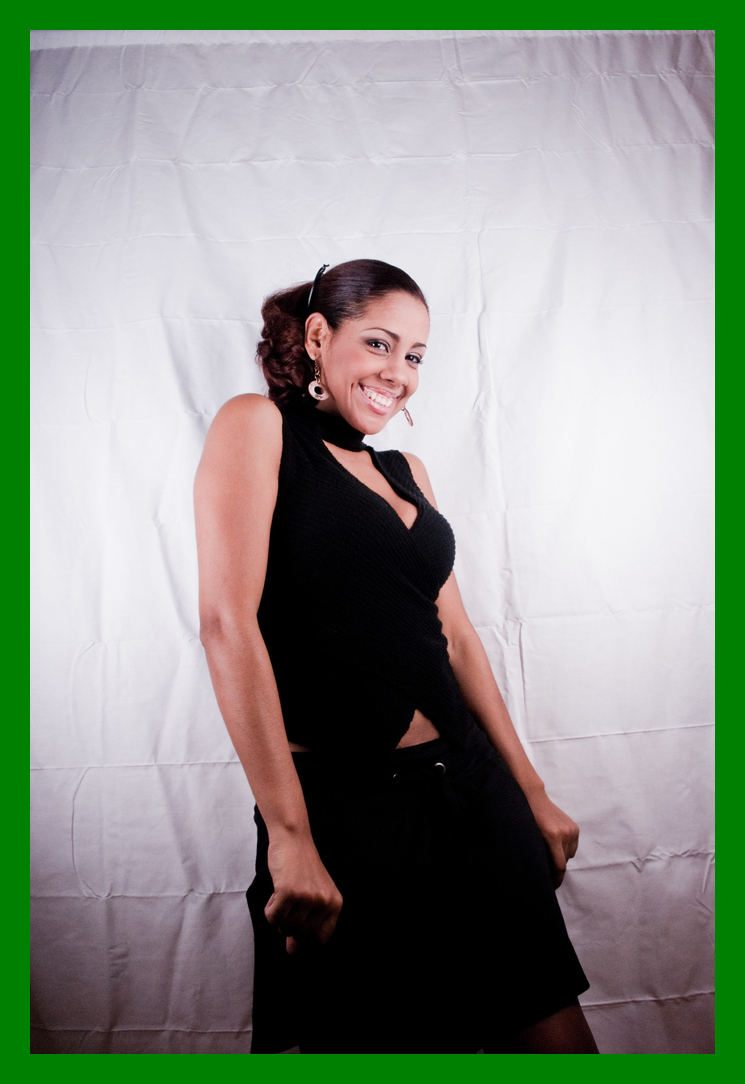}
         \caption{{Image 10}}
     \end{subfigure}
\caption{
Example of \textbf{coreference}:
\textit{``
A woman with a white background smiles at the camera. Most of \textbf{her} body is visible. \textbf{She} is wearing a black outfit.
''}
Definition: Linguistic coreference.
}
\end{figure}

\begin{figure}[t!]
\centering
     \begin{subfigure}[t]{0.49\columnwidth}
         \centering
         \includegraphics[width=\textwidth]{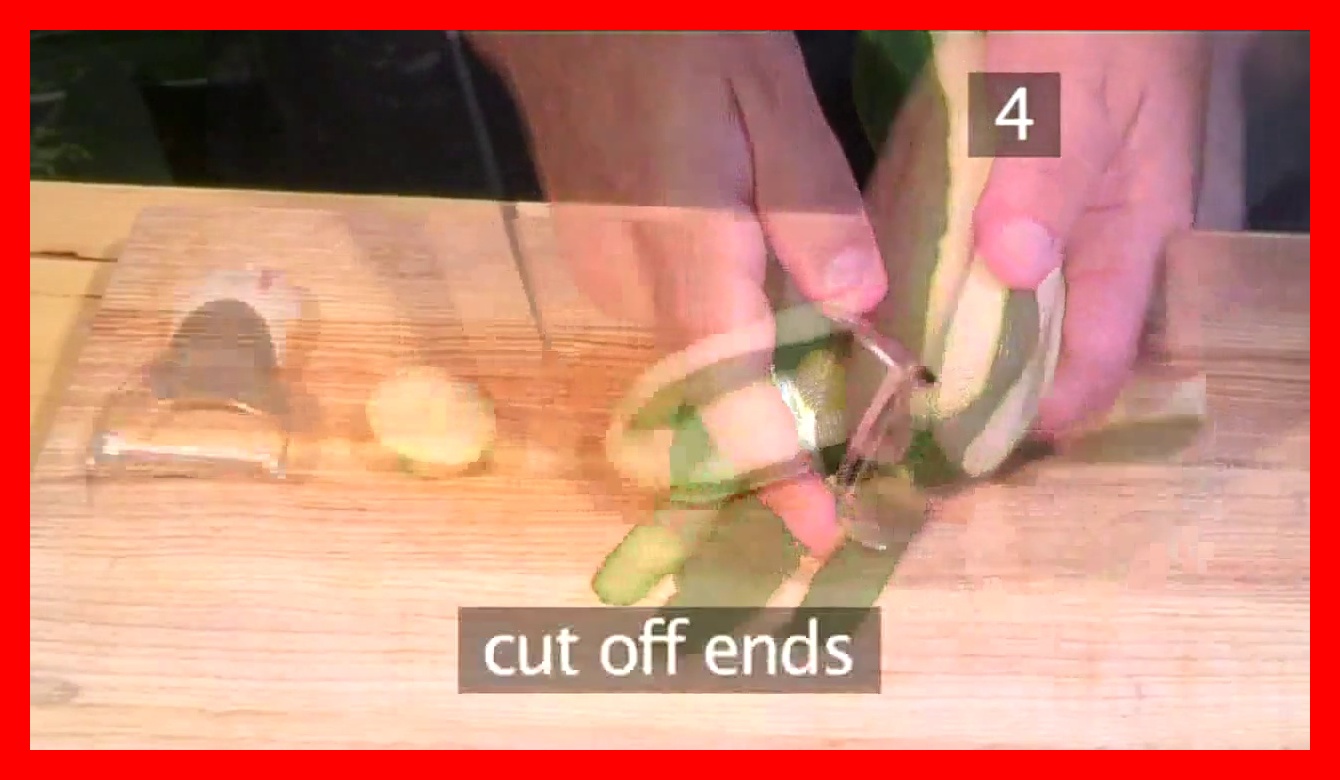}
         \caption{{Frame 7}}
     \end{subfigure}
     \hfill
     \begin{subfigure}[t]{0.49\columnwidth}
         \centering
         \includegraphics[width=\textwidth]{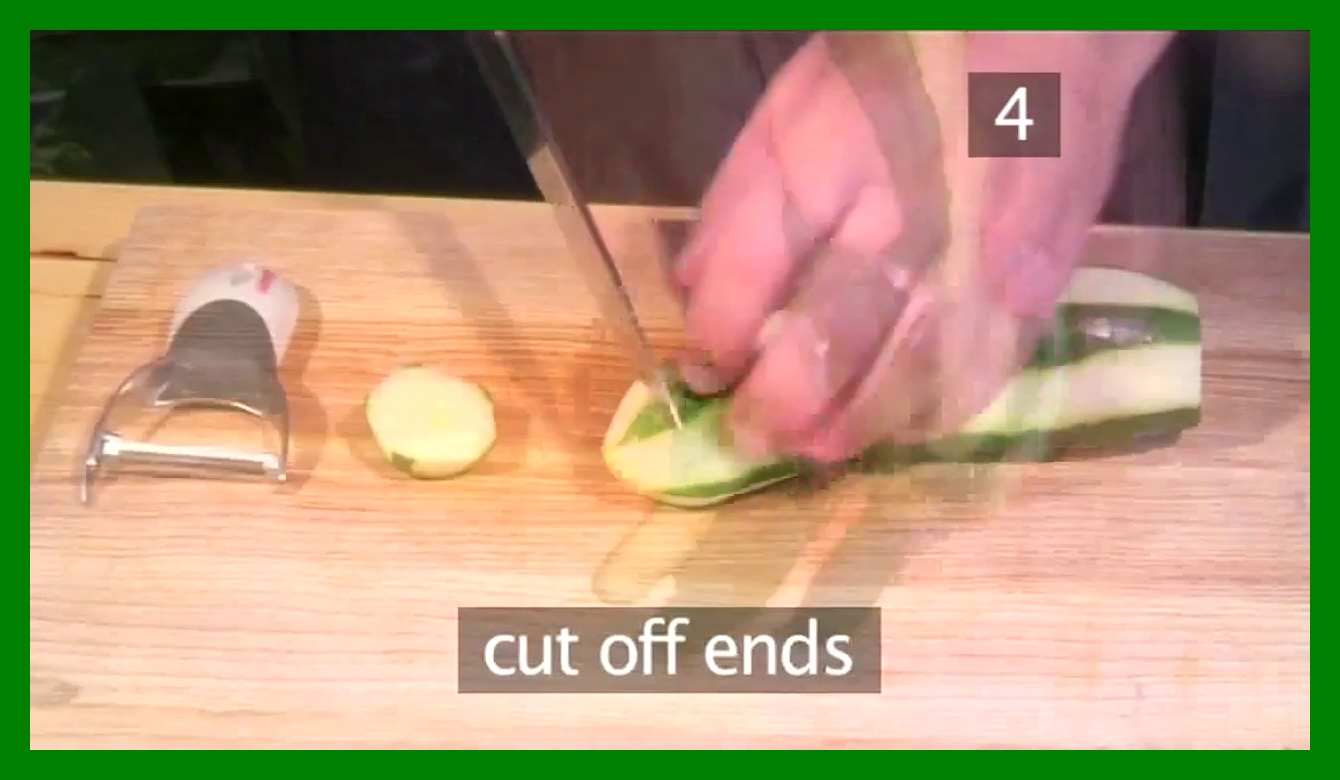}
         \caption{{Frame 8}}
     \end{subfigure}
     
     \begin{subfigure}[t]{0.49\columnwidth}
         \centering
         \includegraphics[width=\textwidth]{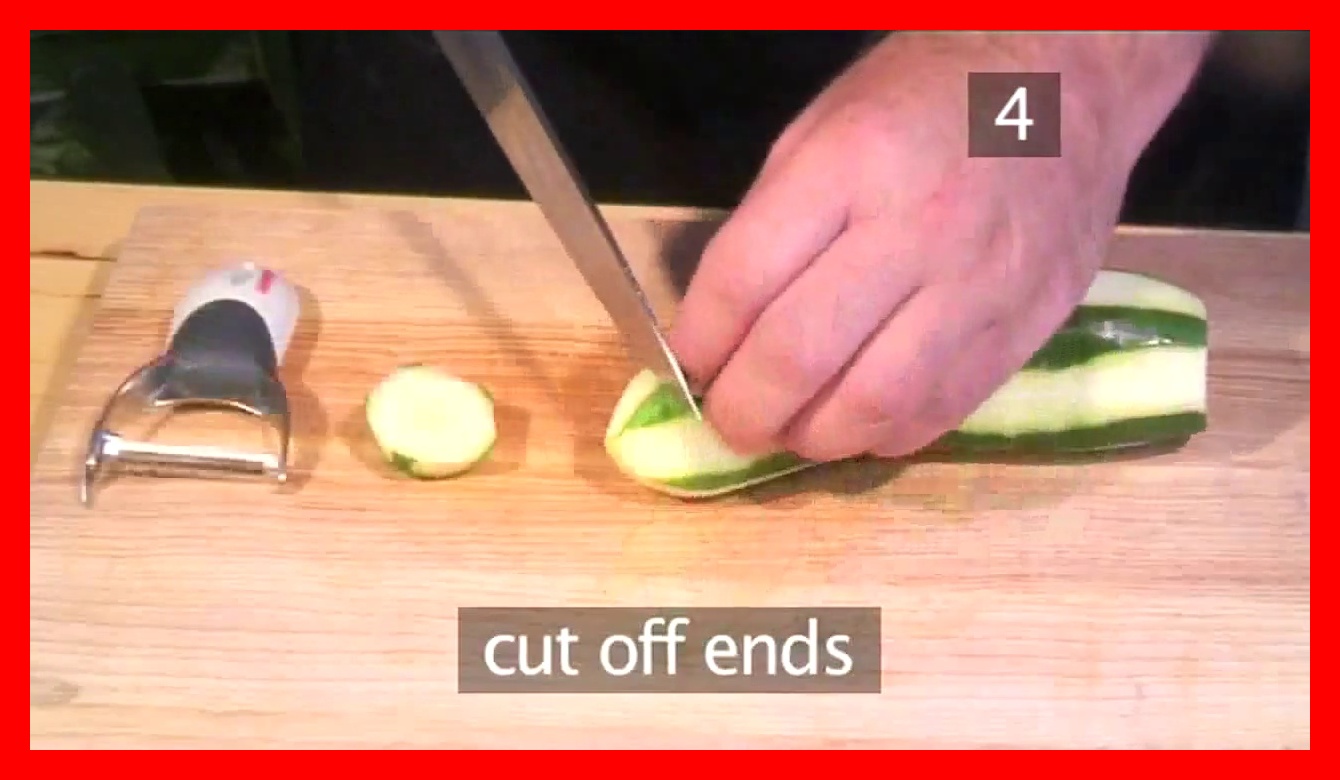}
         \caption{{Frame 9}}
     \end{subfigure}
     \hfill
     \begin{subfigure}[t]{0.49\columnwidth}
         \centering
         \includegraphics[width=\textwidth]{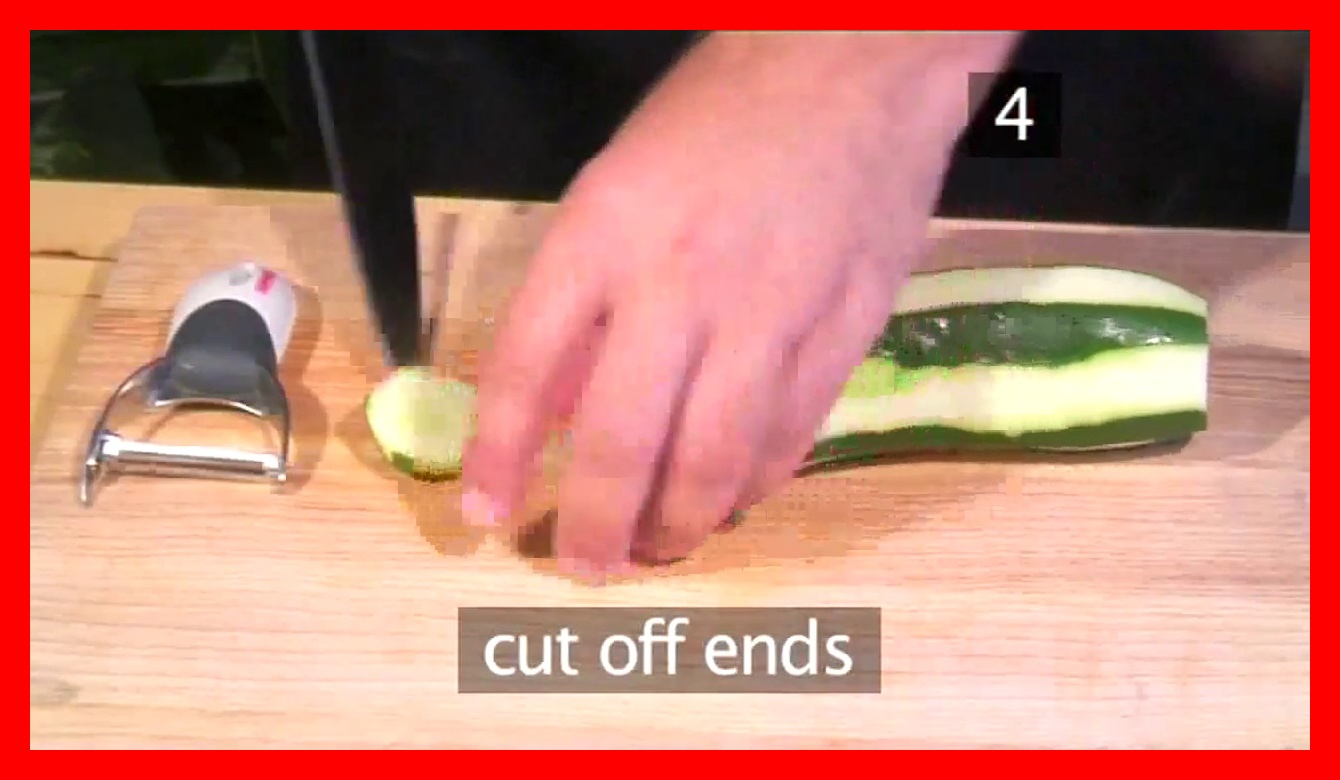}
         \caption{{Frame 10}}
     \end{subfigure}
\caption{
Example of \textbf{meta properties}:
\textit{``
The cucumber is just to be cut into, you can see a \textbf{transparent image covering the image}.''}
Definition: Descriptions that mention aspects that stem from the way the photo/video was taken: two overlayed images (when a video transitions), black-and-white, blurriness, brightness.
}
\end{figure}

\end{document}

%% file: macros.tex
\newcommand{\datasetname}{{Image Retrieval from Contextual Descriptions}}
\newcommand{\datasetacronym}{\textsc{ImageCoDe}}

\newcommand{\contextbatch}{\textsc{+ContextBatch}}
\newcommand{\contextmodule}{\textsc{+ContextModule}}
\newcommand{\temporalemb}{\textsc{+TemporalEmbeddings}}

\newcommand{\xx}{\mathbf{x}}
\newcommand{\mm}{\mathbf{m}}
\newcommand{\vt}{\mathbf{t}}